\documentclass[10pt,twocolumn,letterpaper]{article}

\usepackage{cvpr}
\usepackage{times}
\usepackage{epsfig}
\usepackage{amsmath}
\usepackage{amssymb}
\usepackage{multirow}
\usepackage{bm}
\usepackage{graphicx}
\usepackage{subfigure}
\usepackage{soul}
\usepackage{caption}
\usepackage[toc,page]{appendix}

\usepackage{booktabs}

\newcommand{\minisection}[1]{\vspace{0.04in} \noindent {\bf #1}\ \ }

\DeclareMathOperator*{\argmin}{\arg\!\min}

\usepackage[pagebackref=true,breaklinks=true,letterpaper=true,colorlinks,bookmarks=false]{hyperref}

\cvprfinalcopy 

\def\cvprPaperID{6105} 

\ifcvprfinal\fi

\begin{document}

\title{Semantic Drift Compensation for Class-Incremental Learning}

\author{Lu Yu$^{1,2}$, Bartłomiej Twardowski$^{2}$, Xialei Liu$^{2}$, Luis Herranz$^{2}$, Kai Wang$^{2}$,  \\
Yongmei Cheng$^{1}$, Shangling Jui$^{3}$, Joost van de Weijer$^{2}$\\
$^{1}$ School of Automation, Northwestern Polytechnical University, Xi'an, China \\
$^{2}$ Computer Vision Center, Universitat Autonoma de Barcelona, Barcelona, Spain\\
$^{3}$ Huawei Kirin Solution, Shanghai, China\\
{\tt\small \{luyu,btwardowski,xialei,lherranz,kwang,joost\}@cvc.uab.es}, \\\tt\small{chengym@nwpu.edu.cn, jui.shangling@huawei.com}
}

\maketitle

\begin{abstract}
Class-incremental learning of deep networks sequentially increases the number of classes to be classified. During training, the network has only access to data of one task at a time, where each task contains several classes. In this setting, networks suffer from catastrophic forgetting which refers to the drastic drop in performance on previous tasks.
The vast majority of methods have studied this scenario for classification networks, where for each new task the classification layer of the network must be augmented with additional weights to make room for the newly added classes.
Embedding networks have the advantage that new classes can be naturally included into the network without adding new weights.
Therefore, we study incremental learning for embedding networks. In addition, we propose a new method to estimate the drift, called semantic drift, of features and compensate for it without the need of any exemplars. We approximate the drift of previous tasks based on the drift that is experienced by current task data.
We perform experiments on fine-grained datasets, CIFAR100 and ImageNet-Subset. We demonstrate that embedding networks suffer significantly less from catastrophic forgetting. We outperform existing methods which do not require exemplars and obtain competitive results compared to methods which store exemplars. Furthermore, we show that our proposed SDC when combined with existing methods to prevent forgetting consistently improves results. \footnote{Code available at \url{https://github.com/yulu0724/SDC-IL}.}
\end{abstract}

\section{Introduction}

Future learning machines should be able to adapt to an ever-changing world. They should continuously learn new tasks without forgetting previously learned tasks. Other than the generally applied setup, where training data for all tasks is simultaneously available, in continual learning tasks are learned in a consecutive manner. At each moment the algorithm has only access to the data of a single task. For deep neural networks, one could finetune the network on the data of the latest task. However, in the absence of training data of previous tasks, the network suffers from \emph{catastrophic forgetting}~\cite{mccloskey1989catastrophic}. This refers to a drastic drop in performance on previous tasks. Continual learning studies strategies to mitigate the impact of catastrophic forgetting~\cite{kirkpatrick2017overcoming,li2018learning,rebuffi2017icarl}.

\begin{figure}[tb]
\begin{center}
  \includegraphics[width=0.39\textwidth]{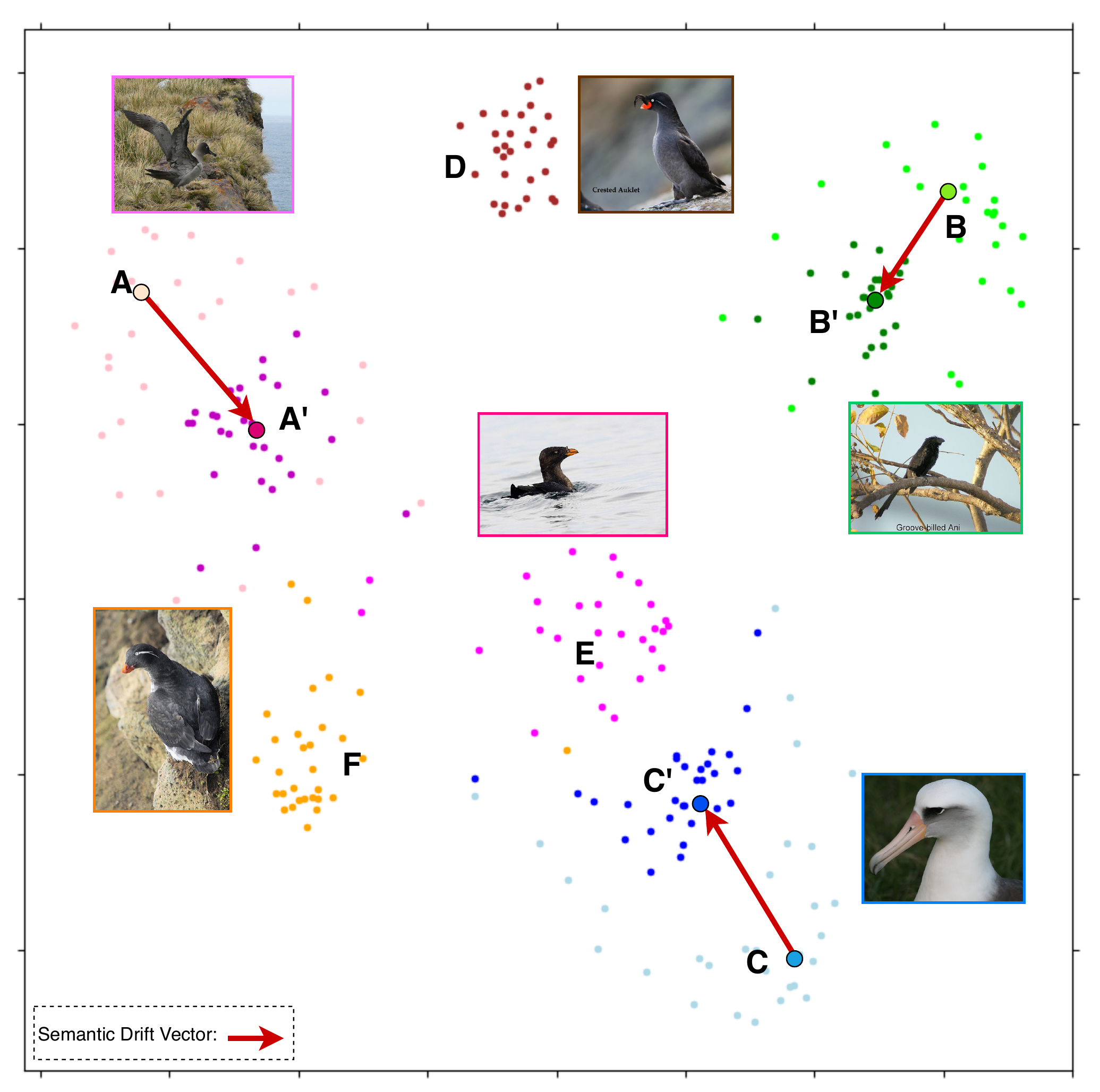}
  \caption{T-SNE visualization of embedding space after finetuning. A, B, C indicate prototypes of task 1 after training task 1; A',B',C' and D,E,F respectively for task 1 and 2 after training for task 2. 
  The semantic drift (indicating forgetting) from task 1 is given by red vectors. Our method estimates this and compensates the prototypes accordingly.}
  \label{fig:tsne}
  \vspace{-2em}
\end{center}
\end{figure}

Continual learning has explored a variety of strategies to prevent networks from forgetting previously learned tasks. Li et al.~\cite{li2018learning} propose a method called learning without forgetting (LwF). They use the same data to supervise learning of the new tasks and to provide unsupervised output guidance on the old tasks to prevent forgetting. Elastic weight consolidation (EWC)~\cite{kirkpatrick2017overcoming} estimates the Fisher matrix to weight a regularization term favouring changes to neurons which were found to be less important in previous tasks, and which prevents the relevant neurons from adapting to the new task. Further research on continual learning includes regularization terms~\cite{ aljundi2018memory,liu2018rotate}, sub-network selection by mask learning~\cite{mallya2018piggyback, mallya2018packnet,serra2018overcoming}, and the use of exemplars~\cite{lopez2017gradient,rebuffi2017icarl}. 

Many of the early works in continual learning considered a task-incremental learning (task-IL) scenario~\cite{van2019three}, in which the network has access to the task-ID at inference time~\cite{aljundi2018memory,kirkpatrick2017overcoming,li2018learning,masana2020ternary,serra2018overcoming}. Recently, more works consider the more difficult class-incremental learning (class-IL)~\cite{aljundi2017expert,dhar2019learning,hou2019learning,liu2018rotate}, where no task-ID is available at inference. The main additional challenge for class-IL is the class imbalance between old versus new tasks. This is addressed by storing data of previous tasks~\cite{Castro2018, hou2019learning,wu2019large}. In this paper, we propose a new method for class-incremental learning. We consider the difficult scenario where no data of previous tasks can be stored. The importance of continual learning algorithms which do not require any storage is growing in a world where data privacy and security are fundamental for many users, and are controlled by government legislation. 

The discussed previous works all study continual learning in classification networks. For these networks, new weights have to be added to accommodate for the newly added classes.
Instead, we perform class-incremental learning for embedding networks which naturally allow for the inclusion of new classes, and do not require network changes for new classes. Embedding networks map data to embedding spaces in which distances correspond to semantic dissimilarities between data points~\cite{chopra2005learning}. They are typically used for image retrieval~\cite{wang2017deep}, face recognition~\cite{schroff2015facenet},
etc. However, they can also be used for classification when combined with, for example, a nearest class mean  classifier~\cite{mensink2013distance}.

In this paper, we show that embedding networks suffer significantly less from catastrophic forgetting than classification networks. We also propose a new method called \emph{semantic drift compensation}. Instead of preventing drift, which most existing methods do, our method estimates the drift of previous tasks during the training of new tasks (see Fig.~\ref{fig:tsne}). We show that an estimate of the semantic drift in previous tasks can be used to compensate for it, thereby improving performance. We evaluate  embedding networks for  image classification by using the nearest class mean (NCM) classifier~\cite{mensink2013distance}. We will refer to the class embedding mean with the term \emph{prototype}. We will show how the drift of prototypes learned in previous tasks can be approximated while only having access to data of the current task. Furthermore, the proposed method can easily be combined with existing methods that prevent forgetting, such as EWC~\cite{kirkpatrick2017overcoming}, LwF~\cite{li2018learning}, or MAS~\cite{aljundi2018memory}, to further improve results.

\section{Related Work}
\minisection{Continual Learning.}  Regularization-based methods optimize network parameters on the current task while preventing the drift of already consolidated weights. Learning without forgetting (LwF)~\cite{li2018learning} adapts a learned model to new tasks while retaining the knowledge gained earlier with a regularization term on probabilities. EWC~\cite{kirkpatrick2017overcoming} and a variant R-EWC~\cite{liu2018rotate} include a regularization term on the weights that forces parameters of the current network to remain close to the parameters trained for the previous tasks. Zenke et al.~\cite{zenke2017continual} propose to compute the consolidation strength of synapses (represented by the network weights) in an online manner, and extends them with a memory to accumulate task-relevant information. Aljundi et al.~\cite{aljundi2018memory} compute the weight importance in a unsupervised manner.

Rehearsal-based methods store a small subset of training data from previous tasks in order to prevent catastrophic forgetting. These exemplars are combined (i.e. rehearsed) with the current task data so that the network parameters are jointly optimized. Some existing works use a distillation loss to prevent forgetting~\cite{chaudhry2018riemannian,lopez2017gradient,rebuffi2017icarl,hou2019learning}. In~\cite{wu2019large}, bias correction is proposed to solve the problem of the data imbalance between the old and new classes especially for large scale datasets. Another alternative is to learn a generative model of previous tasks, and generate synthetic samples (i.e. pseudo-rehearsal) that are combined as usual~\cite{wu2018memory,shin2017continual}. 

We consider continual learning in the challenging class-incremental learning (class-IL) scenario. Some aforementioned methods can be applied directly to class-IL by adjusting the network architecture, e.g.~\cite{aljundi2017expert, rusu2016progressive}. However, they do not scale with the number of classes---as the network requires constant expansion. In~\cite{dhar2019learning} the presented approach uses an attention distillation loss to penalize the changes on attention maps, which helps to retain information of the previous classes, whenever new classes are added. Three strategies to prevent forgetting: cross normalization, less-forgetting constrain and inter-class separation with saved exemplars from previous tasks are introduced in~\cite{hou2019learning}. A method applicable for a class-incremental problem was presented in~\cite{liu2018rotate}, where EWC is combined with a reparametrization of the network in form of a factorized rotation what results in a better performance on previous tasks. The authors of work~\cite{wu2019large} discussed the problems of large number of classes and visual similarities between new versus old classes. They proposed a bias correction of network's outputs for new classes by a linear models. A distillation loss was used in~\cite{Castro2018} with an additional exemplars set in order to prevent forgetting for class-incremental learning. Finally, ~\cite{belouadah2019il2m} exploits a dual memory to reduce the negative effect of catastrophic forgetting in image recognition. A model integrates knowledge distillation and retrospection along with the path selection strategy is proposed to overcome catastrophic forgetting in ~\cite{rajasegaran2019random}.

Our method differs with previous work in two principal aspects. To train new tasks, we use an embedding network with a metric learning loss rather than a classification loss. And secondly, all of the methods discussed above focus on preventing forgetting during the learning of new tasks. Our method does not focus on preventing forgetting, but instead proposes to estimate the drift of features that happens due to the learning of new tasks. Having an approximation of the drift, we can compensate the prototypes of previous tasks. 

\minisection{Deep Metric Learning.} 
Siamese networks~\cite{chopra2005learning} were first proposed to learn embeddings for face verification. Initially, they used contrastive loss, which ensures that pairs from the same category are close and pairs from different categories are far.
Triplet networks~\cite{hoffer2015deep,wang2014learning} were proposed to address the limitations of contrastive loss. The inputs are an anchor image, a positive and a negative image. The aim of a triplet network is to learn embeddings for which the distance between the similar pairs is smaller than the distance between the dissimilar pairs. 
A direct extension to this method is a quadruplet network~\cite{chen2017beyond}, where the number of points is extended to four and three pairs are used at once in a loss function. 
Further improvements to the embedding learning include: constraining the angle at the negative point of triplet triangles~\cite{wang2017deep} and exploiting all selected pairs information in a multi-similarity loss function proposed in \cite{Wang2019}.

The advantages of embedding networks, when compared to classification networks, is an ongoing discussion. Recent works have pointed out serious shortcomings of classification networks, mostly attributed to the cross-entropy loss (which is based on a softmax operation). Embedding networks were found to be more robust to the exposure of adversarial examples, and better in the detection of out-of-distribution examples~\cite{masana2018metric,rippel2016metric}. Furthermore, deep embeddings were reported to be superior to classification networks for transfer learning~\cite{scott2018adapted} and preliminary results suggest that they might be less prone to catastrophic forgetting~\cite{yang2018robust}. 

\section{Continual Learning for Embeddings}
We consider a class-incremental learning setup where a network learns several tasks, each task containing a number of new classes. During the training of task $t$ we only have access to data $D^t$ which contains pairs $\left({\bf x}_{i}, y_i\right)$, where ${\bf x}_{i}$ is an image of class $y_i \in C^t$. For each task we consider that there is data of a limited set of classes $C^t=\{c^t_1,c^t_2...,c^t_{m^t}\}$, where $m^t$  is the number of classes in task $t$. We consider the generally studied case where there is no overlap between the classes of different tasks: $C^t \cap C^s=\varnothing $ for $t \neq s$. After training all $n$ tasks we evaluate the learned embedding on all classes $C=\bigcap_i C^i$. As other class-incremental methods, we consider a \emph{task-agnostic} setting where the algorithm has no access to the task label at test time. 

\begin{figure*}[tb]
\centering
\includegraphics[width=0.31\textwidth]{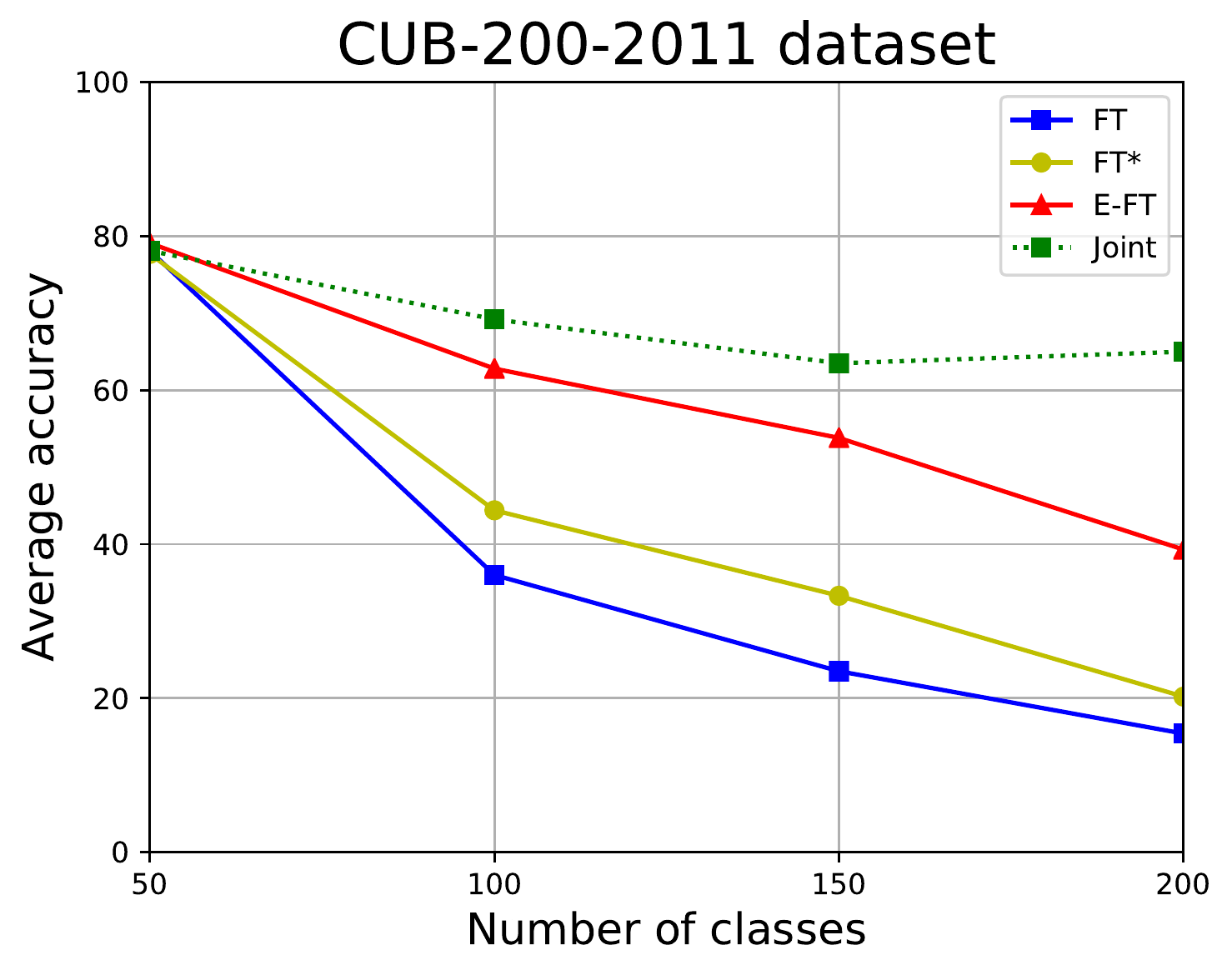}\qquad
\includegraphics[width=0.3\textwidth]{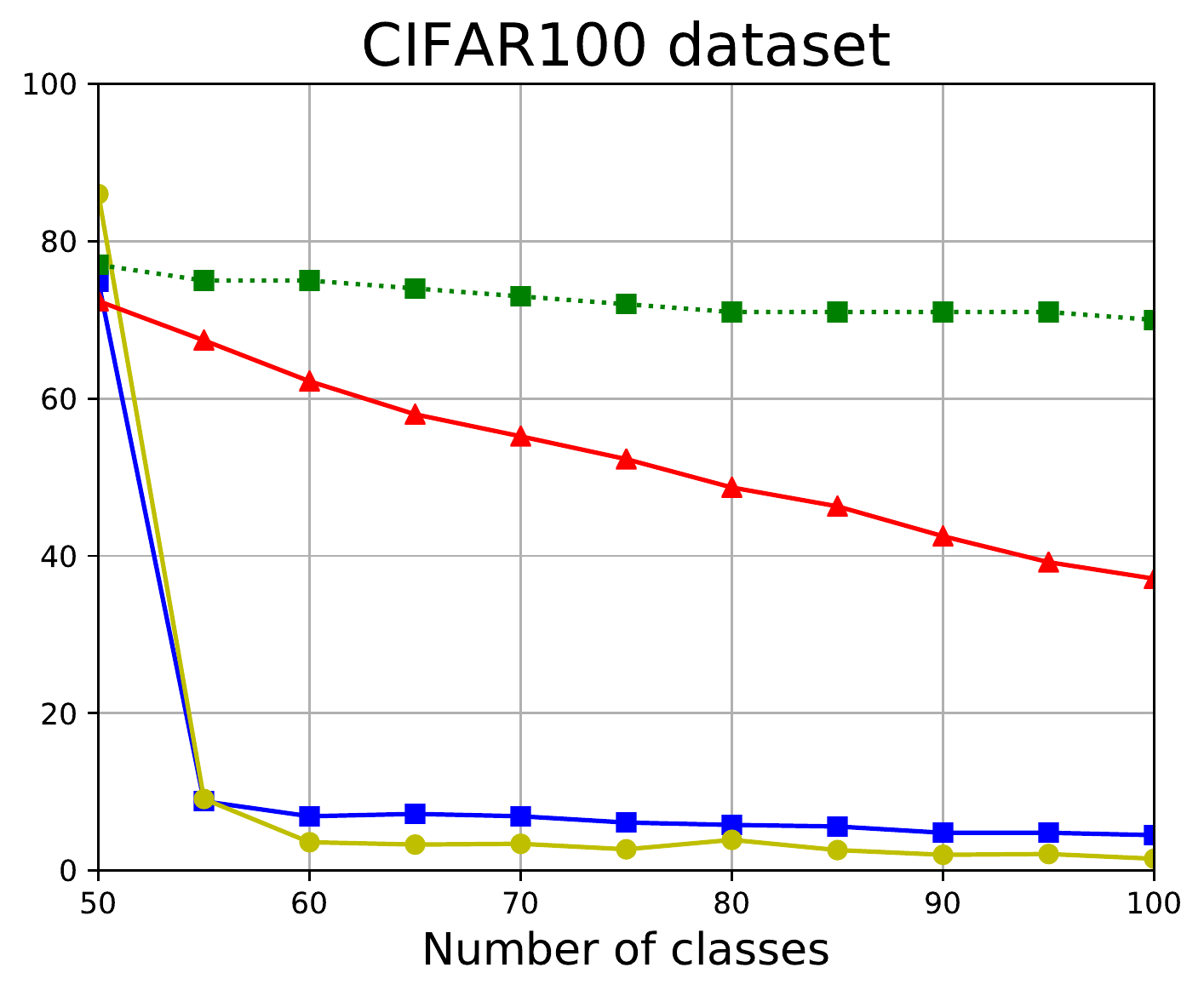}\qquad
\includegraphics[width=0.3\textwidth]{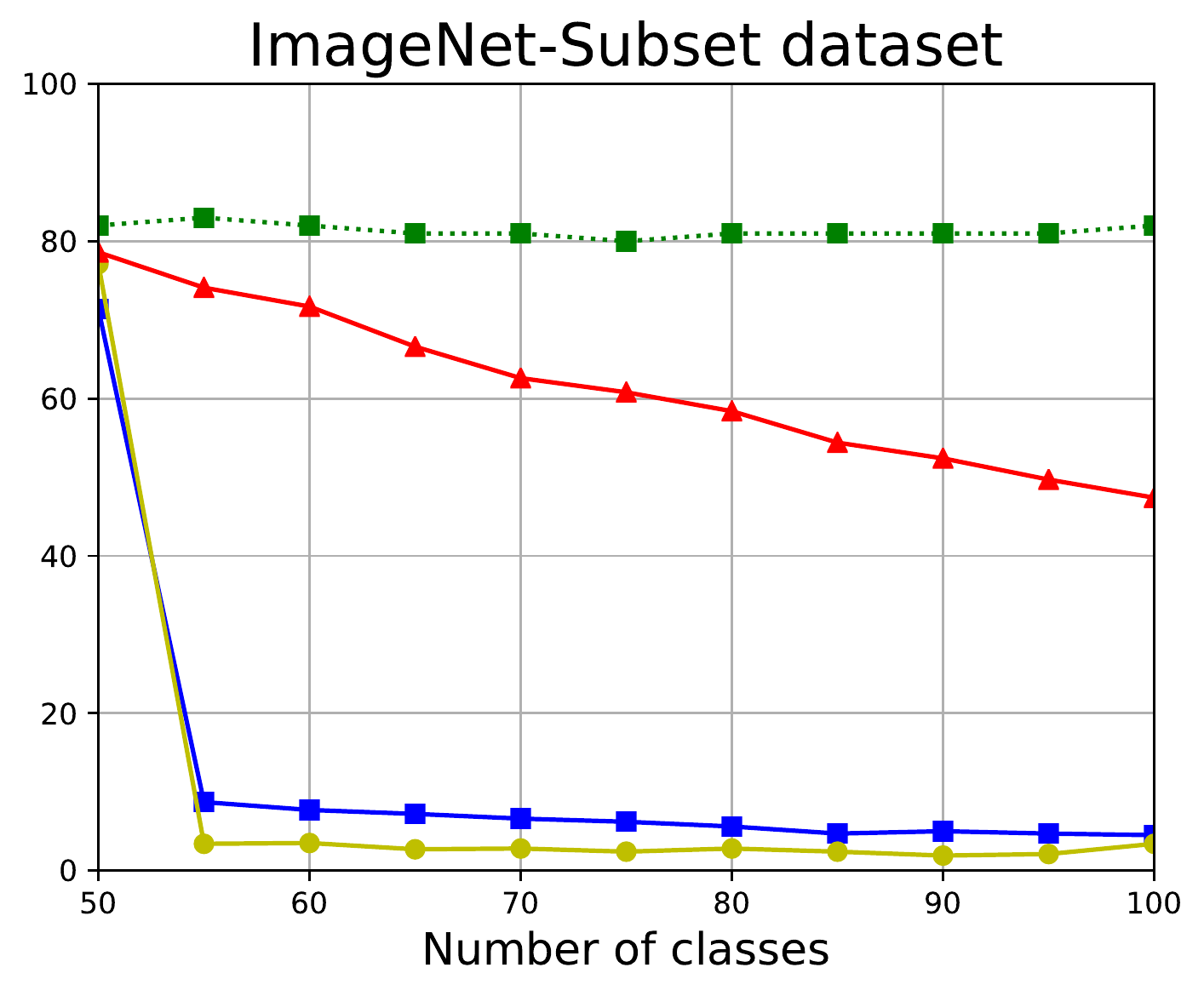}\\
\caption{
Average incremental accuracy for finetuning trained and evaluated with softmax (FT) , finetuning trained with softmax but classified with NCM (FT*), and trained with metric learning loss and classified with NCM (E-FT). In addition, we report joint training (Joint). The results show that continual learning with embedding networks suffers from significantly less forgetting.
}
\vspace{-1\baselineskip}
\label{fig:average accuracy}
\end{figure*}

\subsection{Embedding Networks}\label{sec:emb}
We start by explaining the training of an embedding network for a single task. Embedding networks map data into a low-dimensional output where distance represents the semantic dissimilarity between the images~\cite{bromley1994signature,chopra2005learning}. They simultaneously perform feature extraction as well as  metric learning. In the learned embedding space it is possible to apply a simple metric, such as L2-distance, to determine the similarity between the original images. 

Chopra et al.~\cite{chopra2005learning} proposed to use Siamese networks with the contrastive loss as an objective function. This loss needs related and unrelated pairs of images, and ensures that the distance between related pairs will be low, and the distance between unrelated pairs larger than a margin. For some embeddings it was found that the contrastive loss is hard to train, and other losses have been proposed. The triplet loss is proposed by Hoffer et al.~\cite{hoffer2015deep} based on the work of Wang et al.~\cite{wang2014learning}. 
The objective function forces the negative instance to be further away from the anchor than the positive ones (plus a margin $m$). The triplet loss is given by:
\begin{equation}
\mathcal{L}_{T}= \max(0, d_+-d_- + m),\label{eq:triple}
\end{equation}
where $d_+$ and $d_-$ are the Euclidean distances between the embeddings of the anchor ${\bf z}_{a}$ and the positive instance ${\bf z}_{p}$ and the negative instance ${\bf z}_{n}$ respectively. Here ${\bf z}_i=F({\bf x}_{i})$ is the output embedding for image ${\bf x}_{i}$.

Having trained an embedding network we can use the embedding space for classification. We will use nearest class mean (NCM) classifier which is defined as:
\begin{equation}
c^*_j= \argmin_{c\in C}{\rm dist}\!\left({\bf z}_j,\bm{\mu}_c\right)
\end{equation}
\begin{equation}
\bm{\mu}_c=\frac{1}{n_c}\sum_i \left[y_i=c \right]{\bf z}_i\label{eq:prototype}
\end{equation}
where $n_c$ is the number of training images for class $c$ and $\left [P \right ]=1$ if $P$ is true, and 0 otherwise. We will refer to $\bm{\mu}_c$ as the \emph{prototype} of class $c$. The terminology of prototypes was also used in several works~\cite{snell2017prototypical,yang2018robust} to refer to class representative points in an embedding space. 

\subsection{Softmax Classifier versus Embedding Learning}
\label{sec:softmax_vs_embedding}
The conventional approach to image classification is a softmax classifier trained with a cross-entropy loss. Due to its success, it has been the natural starting point for methods studying continual learning for image classification. However, there are several fundamental drawbacks of the softmax classifier which might limit its application to continual learning. First, network outputs are tightly coupled with prediction classes. Whenever new object classes are added, structural changes to the architecture are required, i.e. new neurons added to accommodate for the new classes. In a class-incremental setting, this results in creating a new output layer (head) for each task~\cite{Castro2018}. Second, in order to acquire a final prediction from a multi-head network, outputs need to be aggregated. Third, the updated model is expected to be biased in the predictions that favor new classes~\cite{hou2019learning, wu2019large}. All of the aforementioned issues can be mitigated to some extent. However, an adaptation process of softmax-based classifiers for class-incremental learning is challenging and harder for longer task sequences. Using embedding networks for continual learning has advantages. New classes can be naturally added without any architectural changes.  While learning new tasks, the network gets fine-tuned to the new data distribution. However, metric learning methods do not require information about classes directly. This is used only to make a proper preparation of the input data, i.e. pairs of positive and negative examples, and thus, the architecture stays intact. 

In order to compare classification and embedding networks in a continual learning setting, we compare both when applying finetuning to adjust to new tasks; a setting which is known to lead to catastrophic forgetting for classification networks. 
The softmax classifier uses new heads for the incremental classification. During testing we compute the probability of each head and take the maximum as the true prediction (called FT). As a second option, we consider performing NCM on the average-pooled output of block 5 of the trained ResNet network, which has the same dimensionality as our embedding network (denoted by FT*). This technique was also used in iCaRL~\cite{rebuffi2017icarl}. The embedding network (trained with triplet loss~\cite{hoffer2015deep}) represents classes with a prototype and performs NCM for classification, and is denoted by E-FT. After convergence on a task is reached, we continue training on a new task with standard stochastic gradient descent, and repeat this until all tasks are learned. 

The results of the comparison on three datasets are presented in Fig.~\ref{fig:average accuracy}. Let ${a_{k,j}} \in [0, 1]$ be the accuracy of the $j$-th task ($j \leq k$) after training the network sequentially for $k$ tasks. Then, the average incremental accuracy at task $k$ is defined as $A_{k}\!=\!\frac{1}{k} \sum_{j=1}^{k}a_{k,j}$~\cite{aljundi2017expert}. We observe that the softmax leads to better classification performance for the first task, as mentioned in~\cite{hoffer2015deep, horiguchi2019significance}. However, the average incremental accuracy of softmax drops significantly when continually adding tasks. This is especially evident for the CIFAR100 and ImageNet-Subset datasets. Using the softmax and classification with NCM on embedding (FT*) suffers similarly from catastrophic forgetting. The finetuning with metric loss (E-FT) significantly better results are obtained, with absolute gains of 23.9\% on CUB-200-2011, with 32.6\% on CIFAR100, and 42.9\% on ImageNet-Subset.
In conclusion, a well-established method for a single-task learning (i.e. softmax) is not optimal when it is used in a continual learning setting. Classification with NMC and embeddings learned with metric losses suffer significantly less from catastrophic forgetting as we observed. Therefore, we propose to use them for continual learning. In the following, we will develop additional methods to further improve continual learning for embedding networks.

\subsection{Regularizing Embedding Networks}

\begin{figure*}[tb]
\begin{center}
  \includegraphics[width=0.9\textwidth]{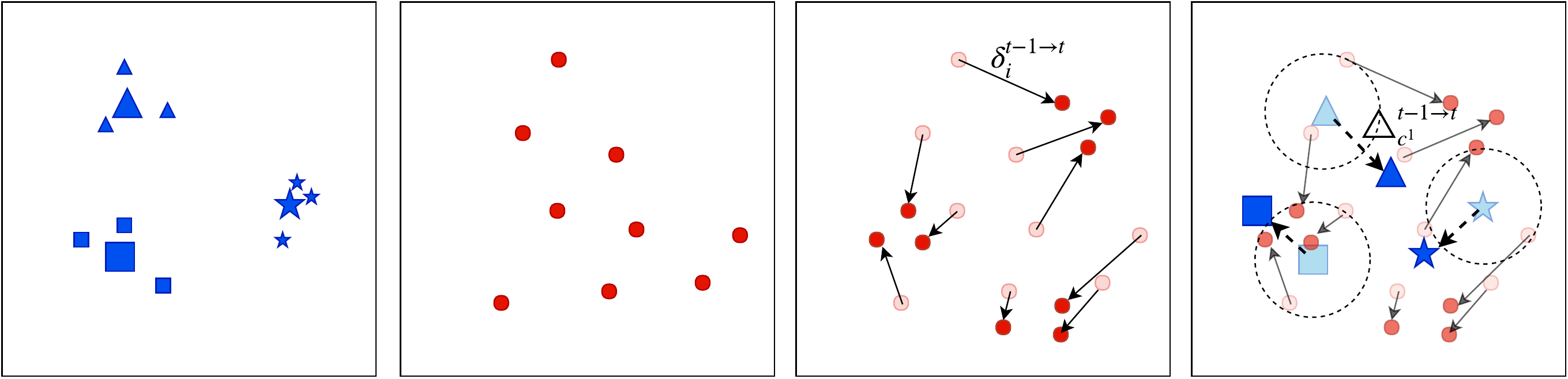}
  \caption{Illustration of semantic drift compensation. (a) Data and prototypes of three classes of task 1 after training task 1. (b) Data of task 2 after training task 1. (c) Drift of data of task 2 while training task 2. This results in a sparse vector field of drift vectors. (d) This vector field is used to approximate the drift of the prototypes of task 1.}
  \vspace{-2em}
  \label{fig:DC}
\end{center}
\end{figure*}

The problem of catastrophic forgetting in continual learning has been extensively studied for classification networks~\cite{chaudhry2018riemannian,kirkpatrick2017overcoming,li2018learning,liu2018rotate,rebuffi2017icarl,shin2017continual,wu2018memory}. To our knowledge, there is no prior work to prevent forgetting the knowledge from the previous tasks on embedding networks. In the following, we adapt several existing techniques to embeddings. We will indicate the variant for embeddings the following notation convention: we append an E (for embedding) to the name of the original method designed for a classification network, e,g, E-LwF would be LwF (Learning without Forgetting) adapted for an embedding network.

\minisection{Finetuning (E-FT)} Described in Sec.~\ref{sec:softmax_vs_embedding} and used as a baseline. For all experiments triplet loss~\cite{hoffer2015deep} is used.

\minisection{Alignment Loss (E-LwF)~\cite{li2018learning}} This method was proposed on classification networks. It aims to match the softmax output of the network of previous models on current data. Instead, on embedding networks, we constrain the parameters drift  by minimizing the distance between the output embeddings of image ${\bf x}_{i}$ during training the current task (${\bf z}^t_i$) with respect to its embedding in the previous task (${\bf z}^{t-1}_i$), similar as~\cite{yu2019learning}. This leads to the following loss:
\begin{equation}
\mathcal{L}_{LwF}=\left \| {\bf z}^t_i-{\bf z}^{t-1}_i \right \|,
\end{equation}
where $\left\| . \right\|$ refers to the Frobenius norm.

\minisection{E-EWC~\cite{kirkpatrick2017overcoming}} This method was proposed on classification networks to keep the network parameters close to the optimal parameters for the previous task while training the current task. This can also be leveraged on embedding networks. The function that we minimize in EWC is:
\begin{equation}
\mathcal{L}_{EWC}=\sum_{p}\frac{1}{2}\mathcal{F}^{t-1}_p(\theta_{p}^t-\theta_{p}^{t-1})^{2},
\end{equation}
where $\mathcal{F}^{t-1}$ is the Fisher information matrix computed after the previous task $t-1$ was learned, and the summation goes over all parameters $\theta_{p}$ of the network.

\minisection{E-MAS~\cite{aljundi2018memory}} This method was proposed to accumulate an importance measure for each parameter of the network based on how sensitive the predicted output function is to a change in this parameter, which can be directly applied to embeddings. The function that we minimize in MAS is: 
\begin{equation}
\mathcal{L}_{MAS}=\sum_{p}\frac{1}{2}\Omega_{p}(\theta_{p}^t-\theta_{p}^{t-1})^{2},
\end{equation}
where $\Omega_{p}$ is estimated by the sensitivity of the squared $l_2$ norm of the function output to their changes.

These losses can be added to the metric learning loss to prevent forgetting while training embeddings continually:
\begin{equation}
\mathcal{L} =\mathcal{L}_{ML} + \gamma \mathcal{L}_{C},
\end{equation}
where $C\in\{LwF,EWC,MAS\}$, $\gamma$ is trade-off between the metric learning loss and the other losses.

\section{Semantic Drift Compensation}\label{sec:SDC}

Embeddings suffer from drift when learned in a sequential manner. When data from previous tasks is not available, using the original prototype in the NCM usually results in a performance drop. We aim at reducing the error that drift causes and propose a drift compensation to update previously computed prototypes. The main idea is to estimate the unknown drift according to the known drift of the current data during the training of the current task.

\subsection{Computation of the Semantic Drift}
\label{sec:CSD}
In Sec.~\ref{sec:emb}, we discussed how prototypes of the classes can be computed for a single task. Here we extend this theory to the continual learning setting. We refer to the prototype mean as $\bm{\mu}_{c^s}^t$ which is the mean for class ${c^s}$ after learning task $t$ computed with Eq.~\ref{eq:prototype}. Class ${c^s}$ is learned during task $s$ (we removed the sub-index $i$ from ${c^s_i}$ for conciseness). When $t>s$ we have no access to data of task $s$ and we cannot compute the true prototype mean (by applying Eq.~\ref{eq:prototype} again). We call the difference between the true class mean and the estimate of the class mean the \textit{semantic drift}:
\begin{equation}
\bm{\triangle}_{c^s}^{s\rightarrow t}=\bm{\mu}_{c^s}^t - \bm{\mu}_{c^s}^s,
\end{equation}
Since we cannot compute $\bm{\mu}_{c^s}^t$ directly we have to find alternative ways to approximate the semantic drift $\triangle_{c^s}^{s\rightarrow t}$. We start by  proposing a method to compute $\bm{\triangle}_{c^s}^{t-1\rightarrow t}$ from which we can then derive $\bm{\triangle}_{c^s}^{s\rightarrow t}$.

When training task $t$ we do not have access to the data of task $s$ and therefore we cannot observe how the embeddings ${\bf z}_i$, for which $y_i \in C^s$, drift during training of task $t$. However, we can measure the drift of the current data  during the training of task $t$. 
\begin{equation}
\bm{\delta}_i^{t-1\rightarrow t}={\bf z}^{t}_i-{\bf z}^{t-1}_i\;, \;y_i\in C^t,
\end{equation}
here we use the notation ${\bf z}^{t}_i$ to refer to the embedding of point $i$ after training task $t$. At the start of training task $t$ we have access to ${\bf z}^{t-1}_i$ which is the embedding of data point $i$ after training task $t-1$.

We propose to approximate the semantic drift $\bm{\triangle}_{c^s}^{t-1\rightarrow t}$ from the sparse vector field $\bm{\delta}_i^{t-1\rightarrow t}$. We do this by interpolating this vector field at the prototype location $\bm{\mu}_{c^s}^{t-1}$ using: 
\begin{equation}
\hat{\bm{\triangle}}_{c^s}^{t-1\rightarrow t}=\frac{\sum_i\left [y_i\in C^t\right] w_i \bm{\delta}_i^{t-1\rightarrow t}}{\sum_i\left [y_i\in C^t\right]w_i} \label{eq:sd},
\end{equation}
with 
\begin{equation}
w_i=e^{-\frac{\left\| {\bf z}^{t-1}_i-\bm{\mu}_{c^s}^{t-1} \right\|^2 }{2\sigma^{2}}}\label{eq:weight},
\end{equation}
where $\sigma$ is the standard deviation of the Gaussian kernel.

In summary, as shown in Fig.~\ref{fig:DC}, for all data points in task $t$ we can monitor the semantic drift during the training of task $t$. This results in a set of drift vectors $\bm{\delta}_i^{t-1\rightarrow t}$ which are used to compute the semantic drift of all previously learned prototypes $\hat{\bm{\mu}}_{c^s}^{t-1}$. This is done by assigning a weight to the drift vectors according to their distance to the prototypes, and computing the prototype drift as a weighted mean of the nearby drift vectors (with Eq.~\ref{eq:sd}).

We can apply the semantic drift compensation (SDC):
\begin{equation}
\hat{\bm{\mu}}_{c^s}^t=\bm{\mu}_{c^s}^s+\hat{\bm{ \triangle}}_{c^s}^{s\rightarrow s+1} + ... +\hat{\bm{ \triangle}}_{c^s}^{t-1\rightarrow t}
\end{equation}
where total compensation is the sum of the compensations which where measured during all previous steps. Normally a recursive scheme would be applied where you update all previously learned prototypes at each new task:
\begin{equation}
\hat{\bm{\mu}}_{c^s}^t=\hat{\bm{\mu}}_{c^s}^{t-1}+\hat{\bm{ \triangle}}_{c^s}^{t-1\rightarrow t}.
\end{equation}

\subsection{Regularized Semantic Drift Compensation}

\begin{figure}[tb]
\centering
\includegraphics[width=0.2\textwidth]{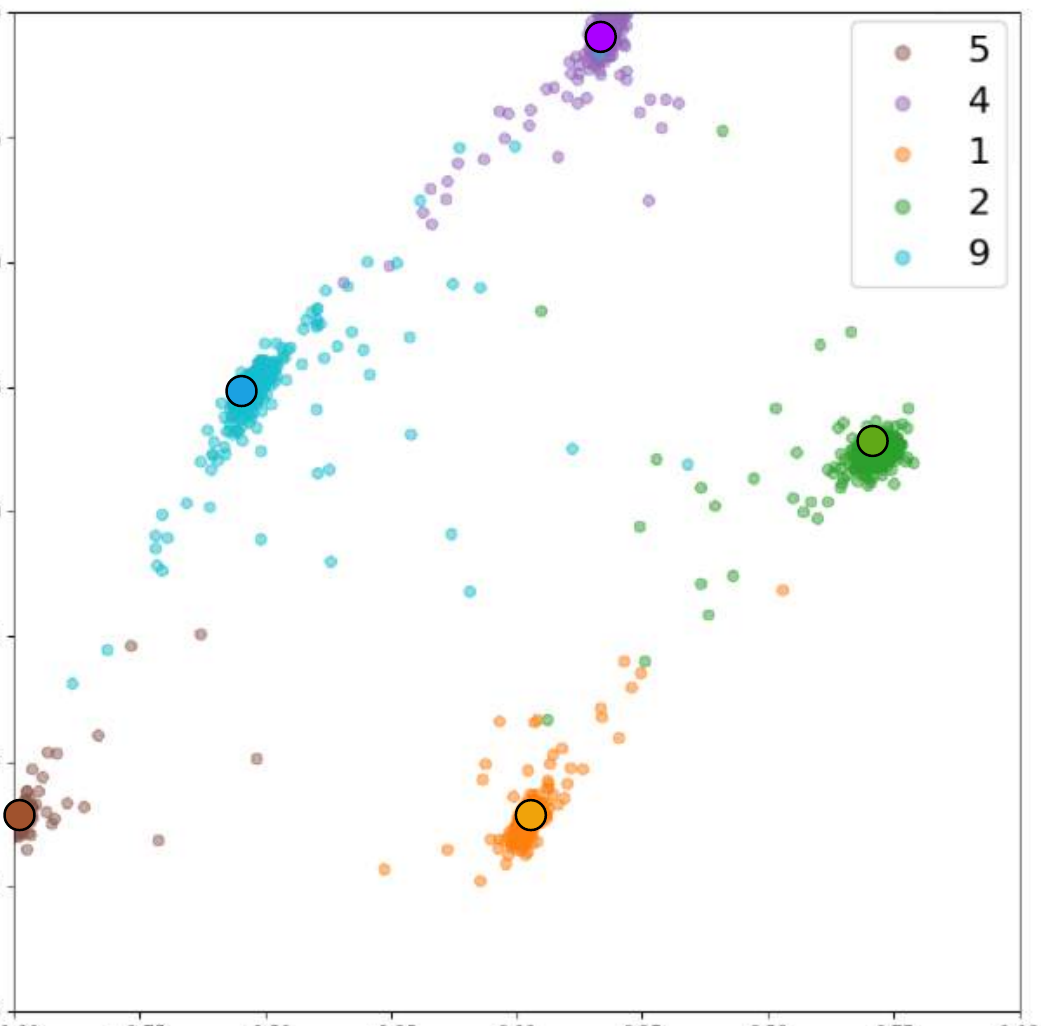}
\includegraphics[width=0.2\textwidth]{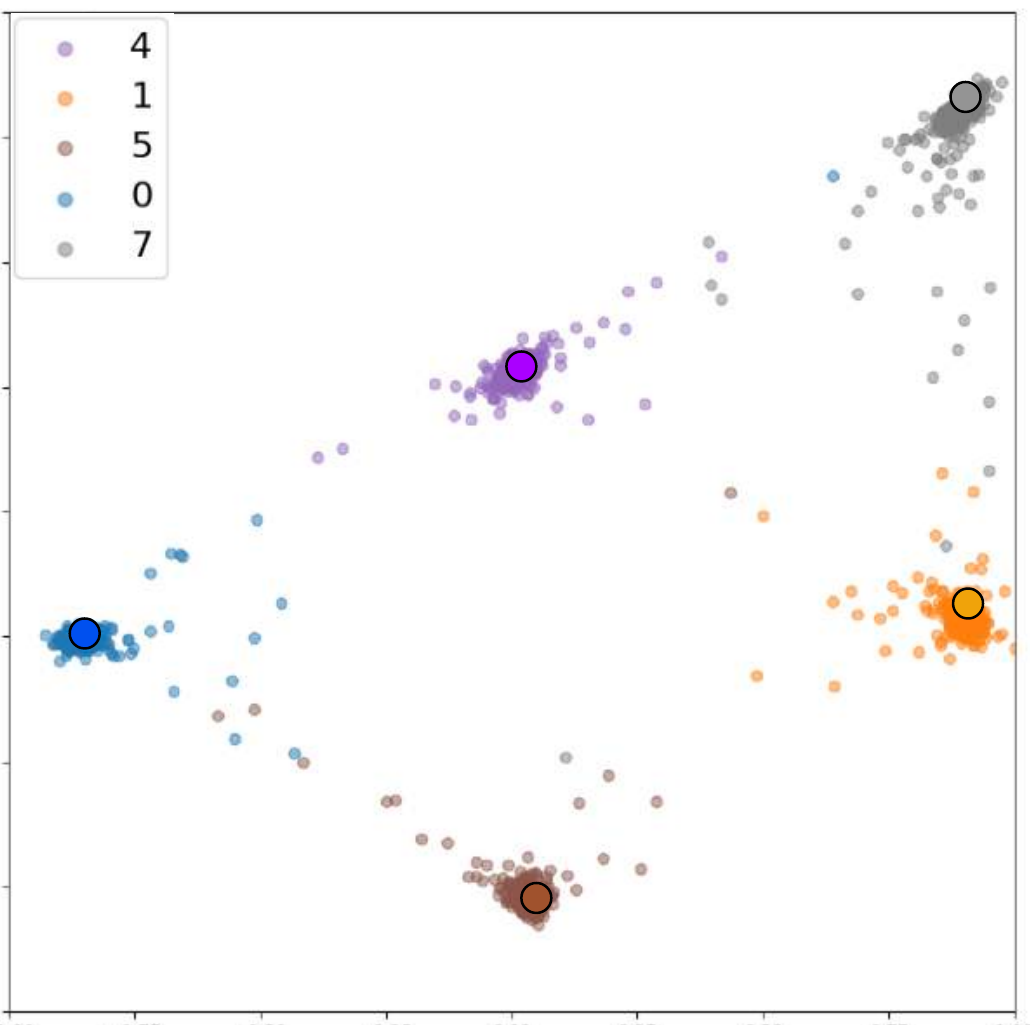} \\
\includegraphics[width=0.2\textwidth]{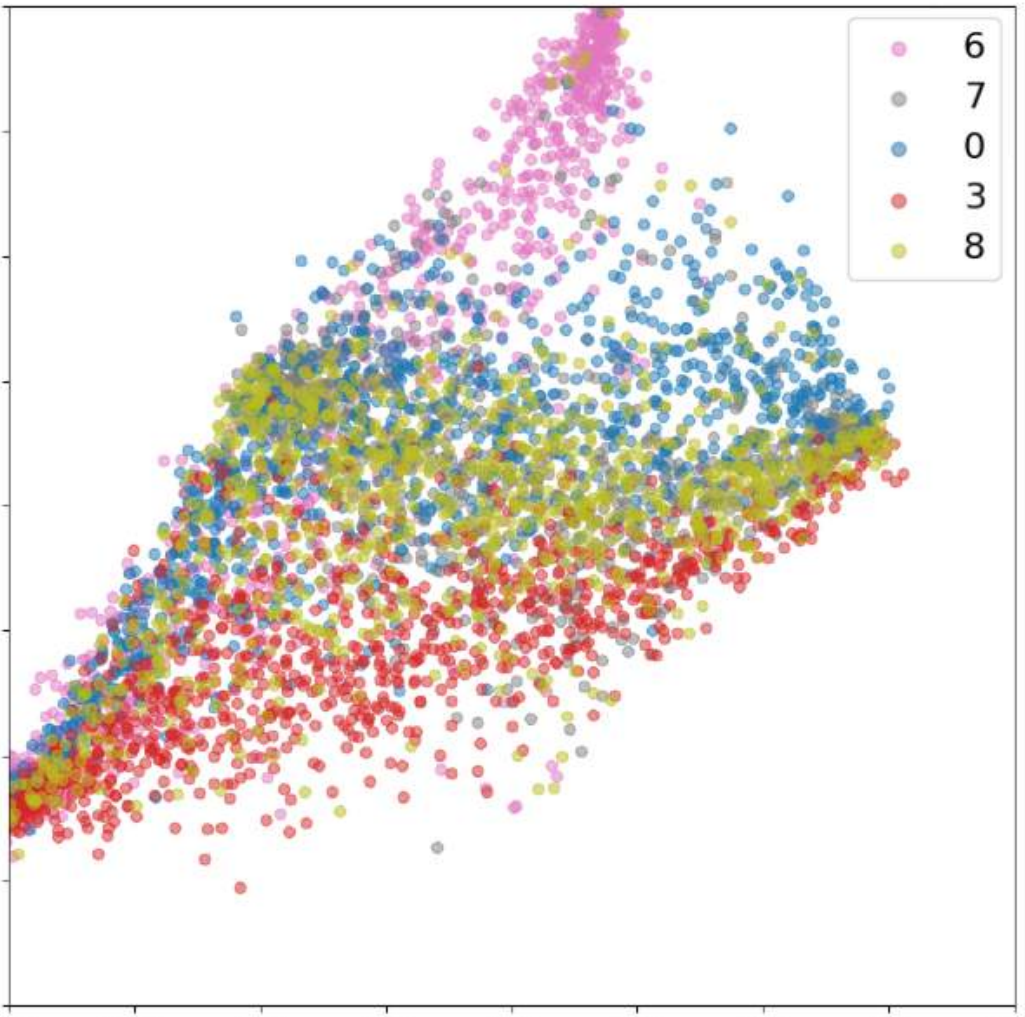}
\includegraphics[width=0.2\textwidth]{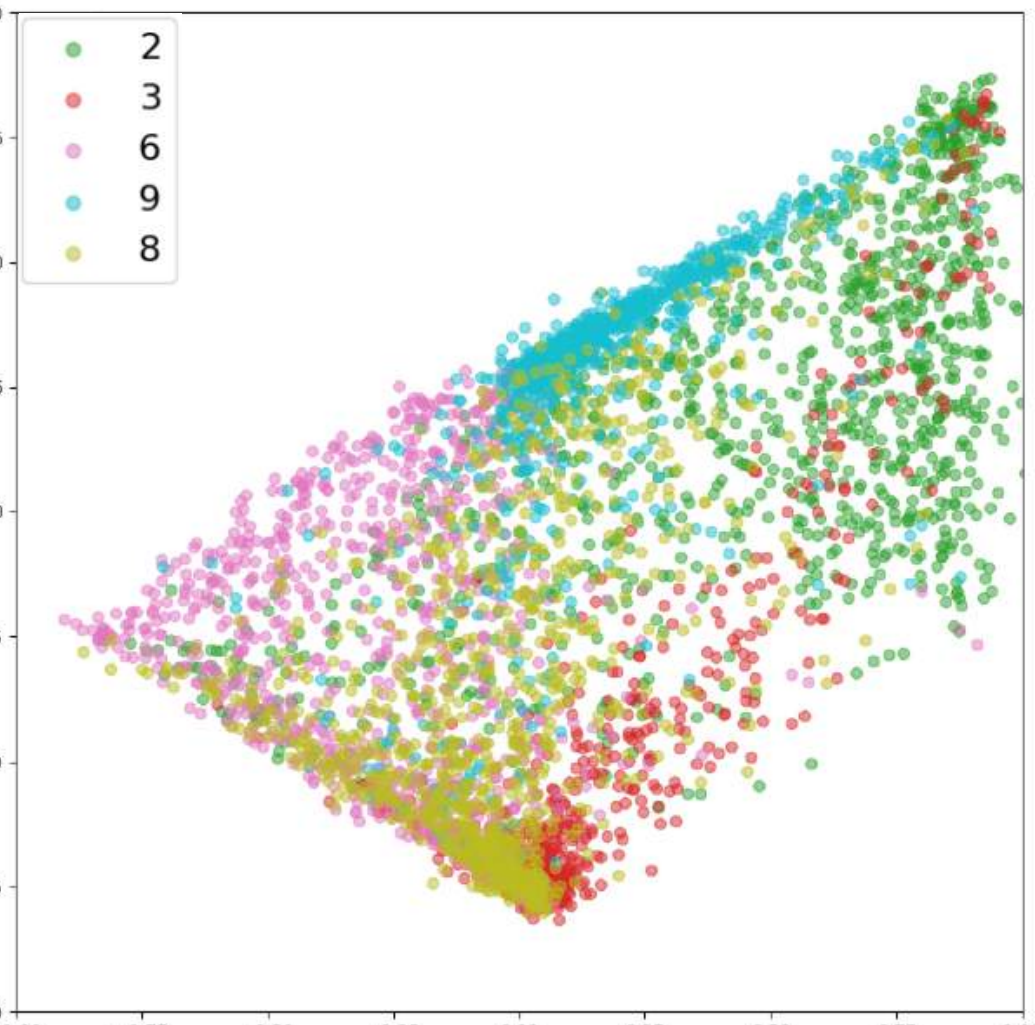} \\
\includegraphics[width=0.2\textwidth]{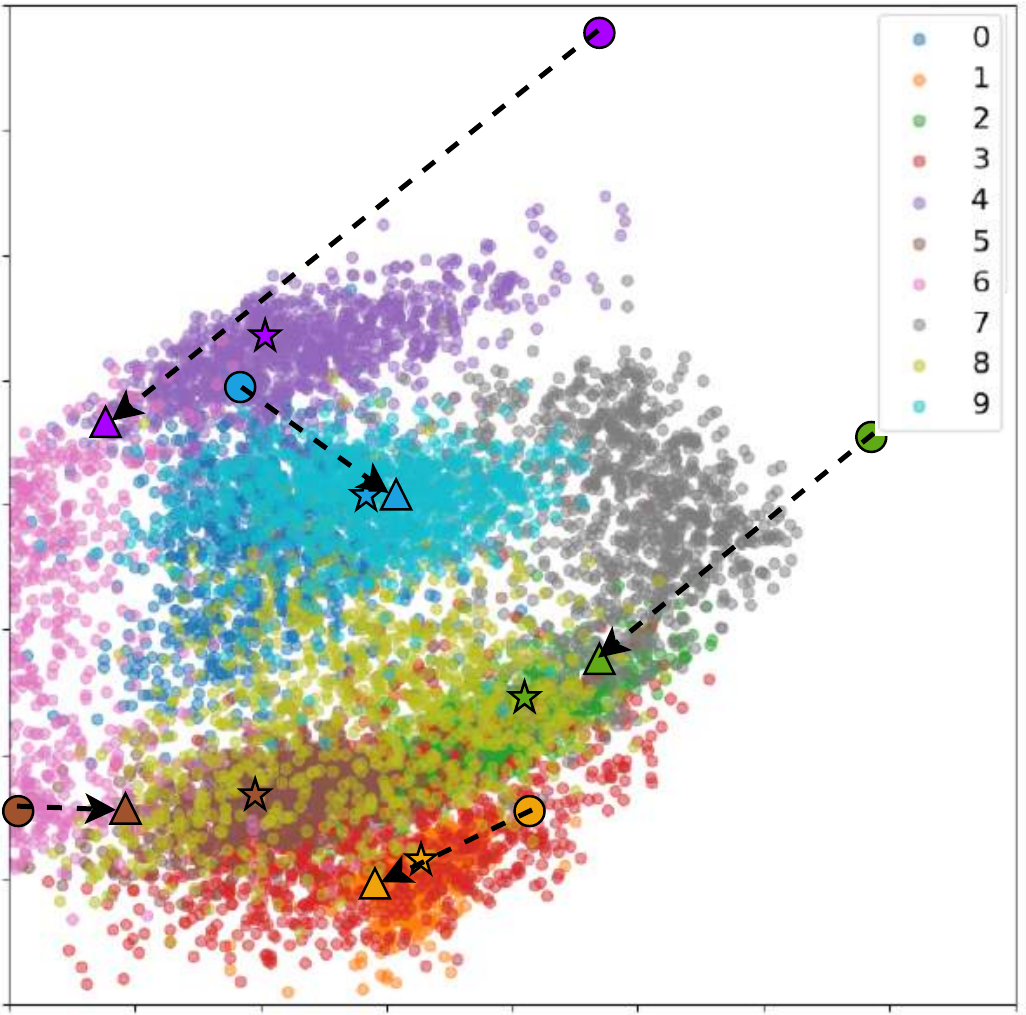}
\includegraphics[width=0.2\textwidth]{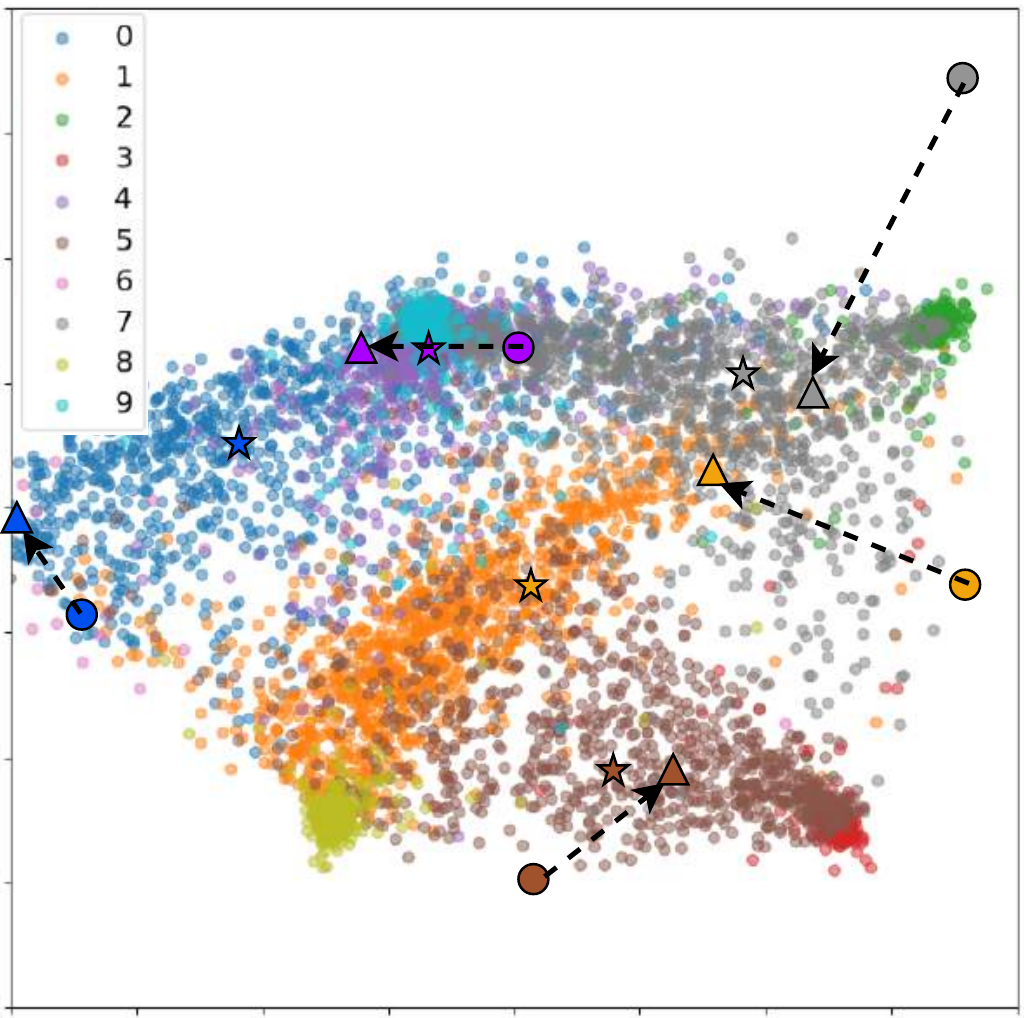} 
\caption{\textbf{Visualization of SDC with E-FT (left) and E-EWC (right).} Top figures represent the embedding of 5 classes of task 1 after training task 1; middle ones represent the embedding of another 5 classes of task 2 after training task 1; bottom ones show the embeddings of both tasks after training task 2. The saved prototypes of the previous task (indicated by circle) are corrected by SDC to new positions (indicated by triangle). Note that the corrected prototypes are closer to the real mean (indicated by star). The dotted arrows are the SDC vectors.} 
\vspace{-1\baselineskip}
\label{fig:real DC}
\end{figure}

Many approaches to continual learning have focused on preventing the network from using parameters which were found to be relevant for previous tasks~\cite{aljundi2018memory,kirkpatrick2017overcoming,li2018learning}. Our method is based on an entirely different approach where we accept the fact that if we share parameters  between the tasks, and we want all tasks to be able to improve (i.e. backpropagate) to all these parameters, this will result in a drift for the previously learned tasks. Approximating this drift allows us then to compensate for it.  Since our approach applies a different methodology to prevent forgetting, it is interesting to see if it is complementary to these other methods. We therefore propose to combine existing methods (E-LwF, E-EWC and E-MAS) with semantic drift compensation and will evaluate this in the experimental results. 

To provide an illustration of SDC, we conduct experiments on MNIST with a 2-dimensional embedding. We divide the ten classes into two disjoint tasks randomly. In  Fig.~\ref{fig:real DC} we show examples of the drift vectors which are estimated by SDC in the case of E-FT and E-EWC\footnote{Examples of the other two methods are in the supplemental material as well as all implementation details and results in tabular form.}. We can see that the approximated drift vectors improve the locations of the prototypes to be closer to the correct positions. As a result, the accuracy of the overall method remains higher while training new tasks.

\section{Experiments}

In this section, we follow the protocol for evaluating incremental learning~\cite{aljundi2018memory,liu2018rotate,rebuffi2017icarl}. For the multi-class datasets, the classes are arranged in a fixed random order. Each method is trained in a class-incremental way on the available data and evaluated on the test set. For the evaluation metric we report: \textit{average incremental accuracy}~\cite{aljundi2017expert} which is the average accuracy of only those classes that have already been trained. We also report \textit{average forgetting}~\cite{chaudhry2018riemannian} on CIFAR100 and ImageNet-Subset dataset. 

\minisection{Datasets.}  We have used the following datasets: CUB-200-2011~\cite{WelinderEtal2010}, 
Flowers-102~\cite{nilsback2008automated}, Caltech-101~\cite{fei2004learning}, CIFAR100~\cite{krizhevsky2009learning}, and ImageNet-Subset containing 100 randomly chosen classes from ImageNet~\cite{deng2009imagenet}. All are divided by classes into tasks randomly. CUB-200-2011 has 200 classes of birds with $11,788$ images in total. Flowers-102 consists of $102$ flower categories of which we randomly choose $100$ with  $8189$ images in total. CIFAR100  contains 600 images for each class. ImageNet-Subset has $129,156$ images in total. Caltech-101 composes of images of objects belonging to 101 widely varied categories.

\minisection{Implementation Details.} All models are implemented with Pytorch. Adam~\cite{kingma2014adam} is used for the optimization. ResNet-18~\cite{he2016deep} is adopted as the backbone network pretrained from ImageNet for CUB-200-2011\footnote{Results on CUB-200-2011 pretrained from ImageNet without birds do not change much, as shown in the supplemental material.} and Flowers-102. For CIFAR100 and ImageNet-Subset version of ResNet-32 and ResNet-18 were used respectively, as in \cite{hou2019learning}, but without pre-training. A triplet loss~\cite{hoffer2015deep} is used in all reported experiments\footnote{The results of using Multi-similarity~\cite{Wang2019} and Angular~\cite{wang2017deep} loss functions are in the supplementary material. Multi-similarity loss improves performance on the first task, but obtain similar results for longer sequences.}. The training images (all resized to $256\times256$, except for CIFAR100 to $32\times32$) are randomly cropped and flipped. We use a mini-batch size of 32. We train our models with learning rate $1\mathrm{e}{-5}$ for $50$ epochs on CUB-200-2011, $1\mathrm{e}{-4}$ for $20$ on Flowers-102, and $1\mathrm{e}{-6}$  for $50$ on CIFAR100 and ImageNet-Subset. The final embeddings of $512$ dimensions are normalized. The trade-off between the E-LwF, E-EWC, E-MAS and triplet loss is $1$, $1e7$ and $1e6$ respectively. We choose a fixed $\sigma=0.3$ to compute the weights of the SDC vectors for all datasets, except for CIFAR100 we choose $\sigma=0.2$.

\begin{table}[tb]
        \setlength\tabcolsep{4.5pt}
        \centering
        \caption{Average incremental accuracy for fine-grained datasets.}\label{tab:main}
        \resizebox{\linewidth}{!}{%
        \begin{tabular}{c|cccccc|cccccc}
        \hline
        \multicolumn{1}{l|}{} & \multicolumn{6}{c|}{CUB-200-2011} & 
        \multicolumn{6}{c}{Flowers-102} \\
        \hline
         & T1 & T2 & T3 & T4 & T5 & T6 
         & T1 & T2 & T3 & T4 & T5 & T6\\
        \hline
        \footnotesize{E-Pre} & 78.5 & 69.1 & 62.1 & 58.1 & 54.7 & 52.1
        &90.9 & 77.5 & 77.7 & 76.1& 75.2&73.6 \\
        \footnotesize{E-Fix} & 84.1 & 70.6 &61.7  &56.9  & 53.5 & 50.3 
        &98.2 & 83.6 & 82.8 &80.1 & 78.4& 76.9\\
        \hline
        \footnotesize{FT} & 79.7 & 34.7 & 23.3 & 17.5 & 12.6 & 11.4 
        & 99.1 & 43.9 & 32.2 & 24.2 & 18.8 &15.3 \\
        \footnotesize{E-FT} & 84.1 & 73.6 & 62.5 & 54.2 & 43.0 & 37.4 
        &98.2 &76.0  & 59.3 & 50.2 & 42.4 &29.1\\
        \footnotesize{E-FT+SDC} & 84.1 & \textbf{75.5} &\textbf{69.5} & \textbf{63.6} & \textbf{57.5} & \textbf{49.3} 
        & 98.2 & \textbf{85.5} & \textbf{74.1} & \textbf{61.9} & \textbf{49.8} & \textbf{35.3}\\
        \hline
        \footnotesize{LwF} & 79.7 & 54.8 & 40.8 & 33.7 & 27.0 & 23.6 
        & 99.1 & 69.7 & 67.4 & 60.0 &49.9 & 46.6\\
        \footnotesize{E-LwF} & 84.1 & 74.0 & 64.8 & 60.0 & 55.5 & 51.4  
        & 98.2 & 85.3 & 81.6 &77.2&69.3& 63.5 \\
        \footnotesize{E-LwF+SDC} & 84.1 & \textbf{74.4} & \textbf{65.9} & \textbf{61.3} & \textbf{57.3} & \textbf{52.7} 
        & 98.2 & \textbf{86.1} & \textbf{82.2} & \textbf{79.6} &\textbf{74.7} & \textbf{69.7}\\
        \hline
        \footnotesize{EWC} & 79.7 & 43.4  & 26.6 & 20.0  & 15.5 & 12.6 
        & 99.1 & 65.2 & 40.9 & 33.8 &23.7 & 22.1\\
        \footnotesize{E-EWC} & 84.1 & 73.6 & 65.0 & 61.6 & 55.0 & 54.2 
        & 98.2 & 86.2 & 84.9 & 82.9 & 80.9 & 79.6\\
        \footnotesize{E-EWC+SDC} & 84.1 & \textbf{74.8} & \textbf{67.4} & \textbf{62.8} & \textbf{58.2} & \textbf{56.4} 
        &98.2& \textbf{87.6} & \textbf{86.9} & \textbf{86.0} & \textbf{84.2} &\textbf{83.9} \\
        \hline
        \footnotesize{MAS} & 79.7 & 49.4 & 37.8 & 31.4 & 25.0 & 22.3 
        & 99.1 & 71.1 & 61.3 & 57.9 &52.1 & 44.8\\
        \footnotesize{E-MAS} & 84.1 & \textbf{72.5} & 65.1 & 60.4 & 54.7 & 51.9 
        & 98.2 & 82.9 & 79.1 & 76.6 & 73.9 & 70.9\\
        \footnotesize{E-MAS+SDC} & 84.1 & 71.9 & \textbf{65.3} & \textbf{61.1} & \textbf{57.3} & \textbf{54.4}  
        & 98.2 & \textbf{83.1} & \textbf{80.7} & \textbf{78.8} & \textbf{76.8} & \textbf{76.0}\\
        \hline
        \end{tabular}}
    \end{table}
    
    \begin{figure}[tb]
        \centering
          \hspace*{-4mm}
          \includegraphics[width=0.5\textwidth]{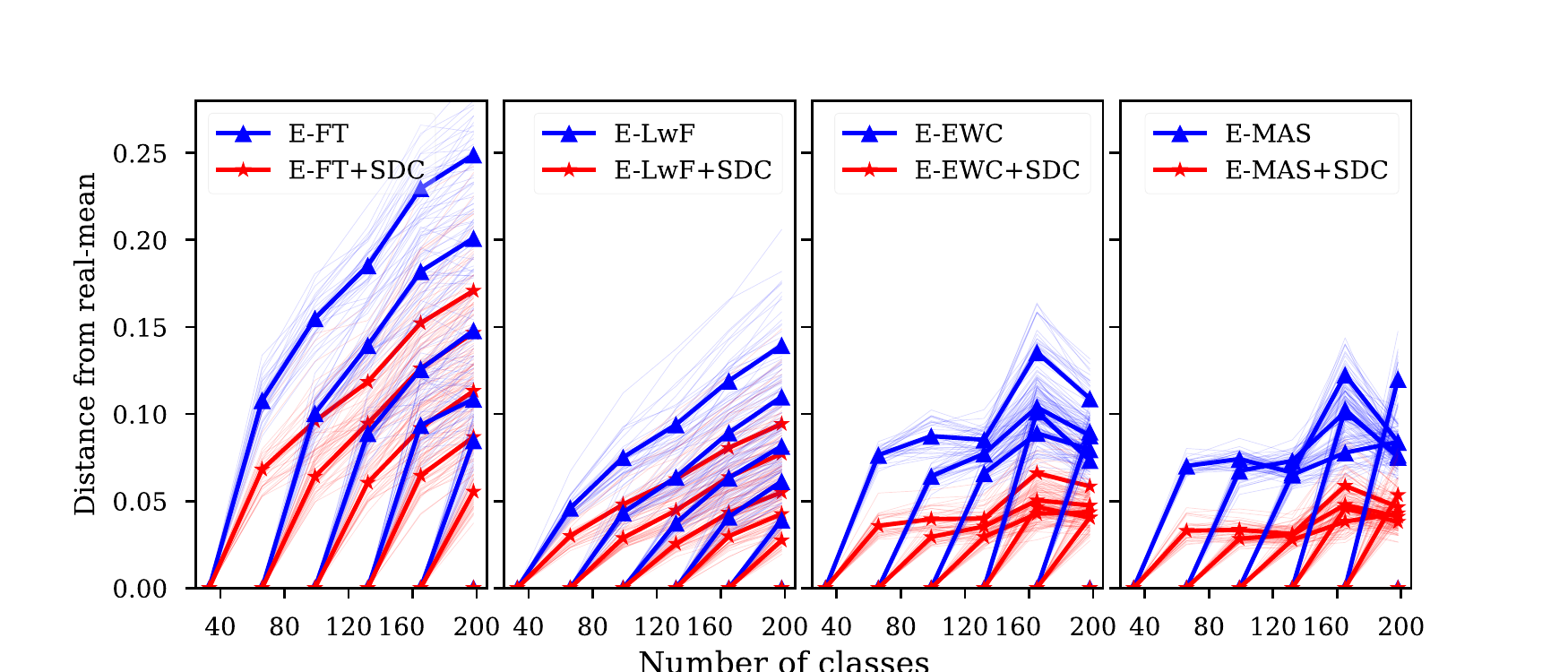}
          \caption{Impact of SDC on the distance between real-mean and prototypes for CUB-200-2011 dataset over tasks. Each line represents a single class. Bold lines represent the mean value of all classes. The graph confirms that SDC correctly compensates for part of the drift of the prototypes.}
          \label{fig:cub-dist-plot}
    \end{figure}

\subsection{Classification with Embedding Networks}

\begin{figure}[tb]
\begin{center}
    \includegraphics[width=0.235\textwidth]{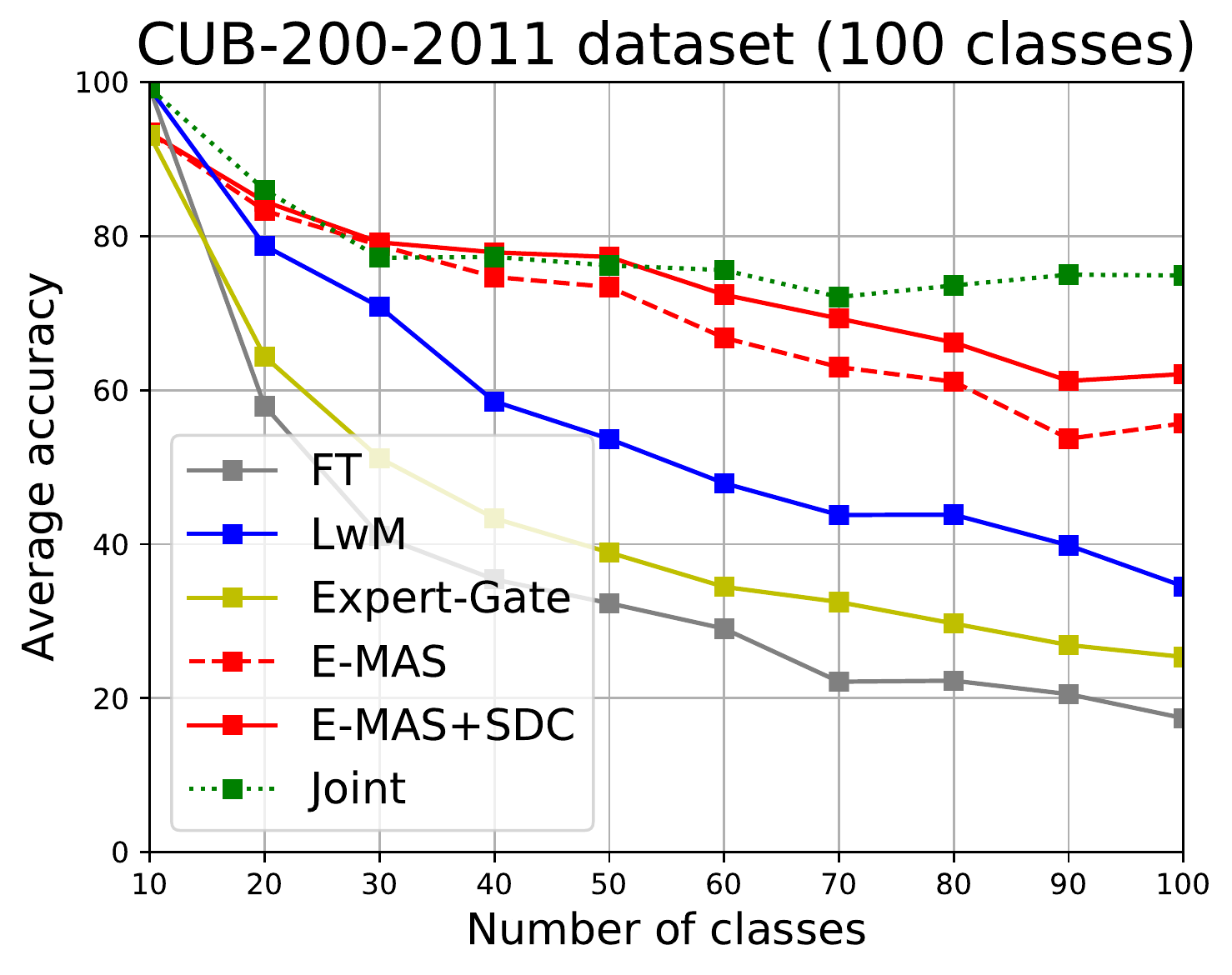}
    \includegraphics[width=0.235\textwidth]{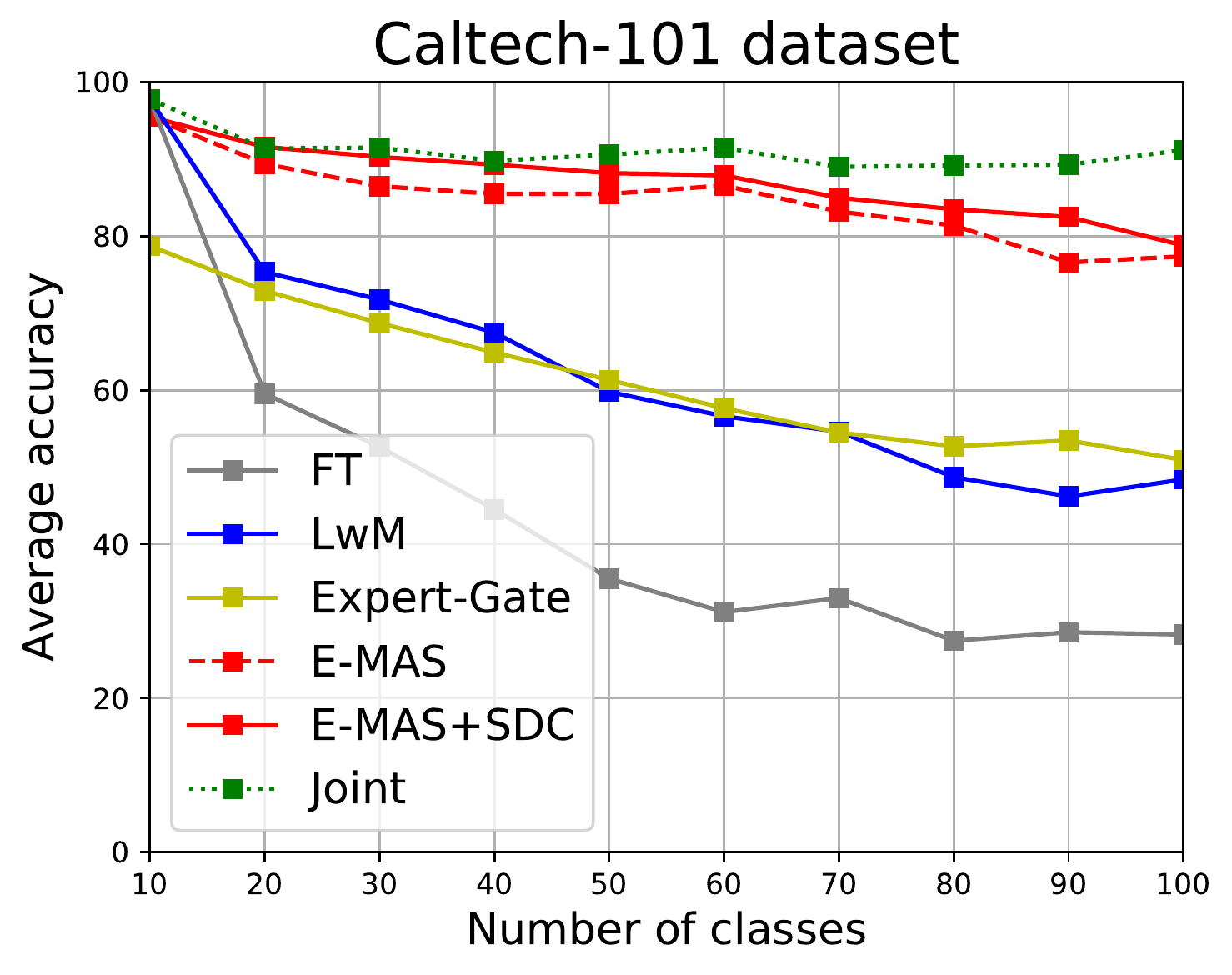}
  \caption{Average incremental accuracy. Comparison of ten-task on CUB-200-2011 (100 classes) and Caltech-101. }
  \label{fig:caltech}
  \vspace{-1\baselineskip}
\end{center}
\end{figure}

\begin{figure*}[tb]
\begin{center}
\includegraphics[width=0.246\textwidth]{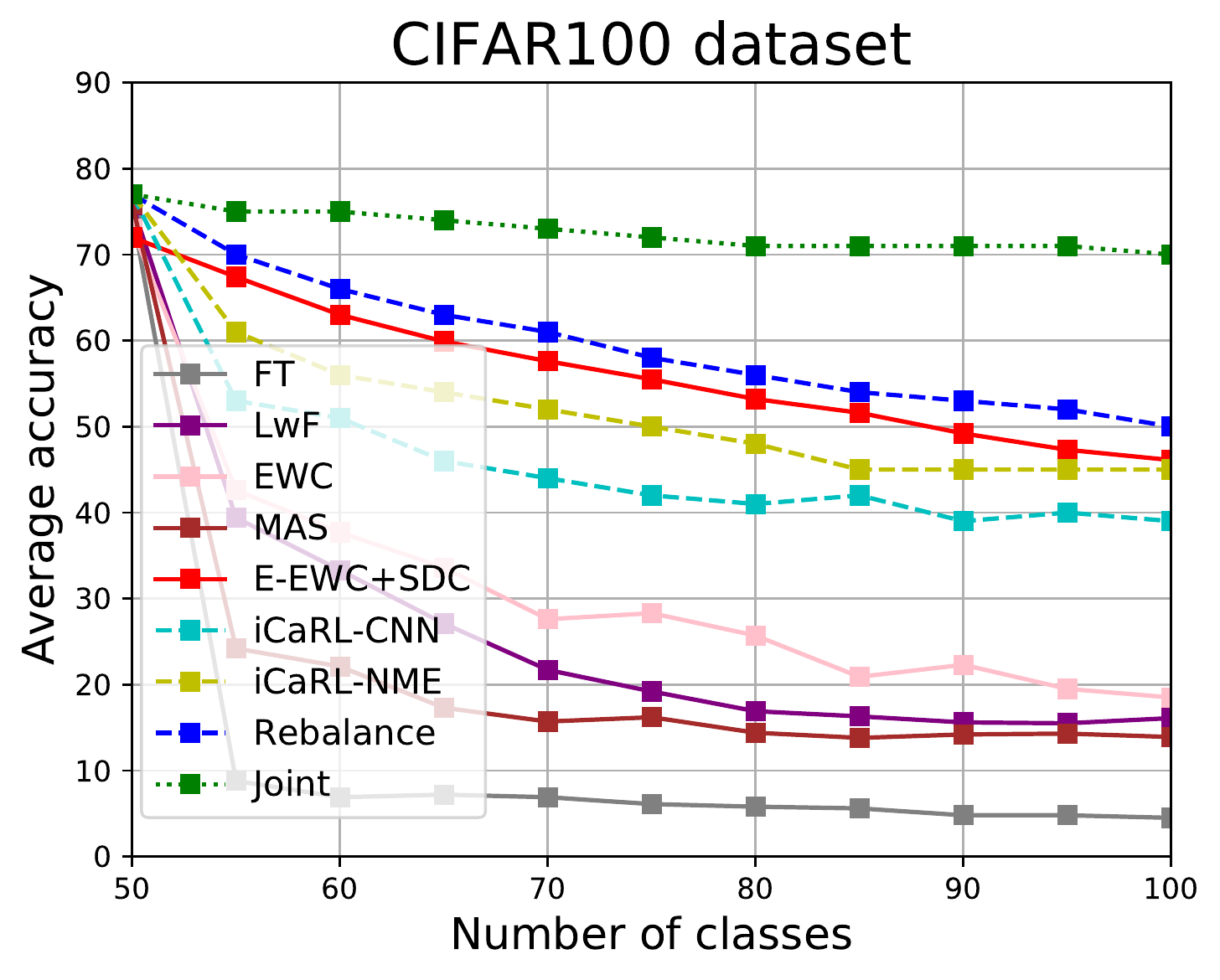}
\includegraphics[width=0.246\textwidth]{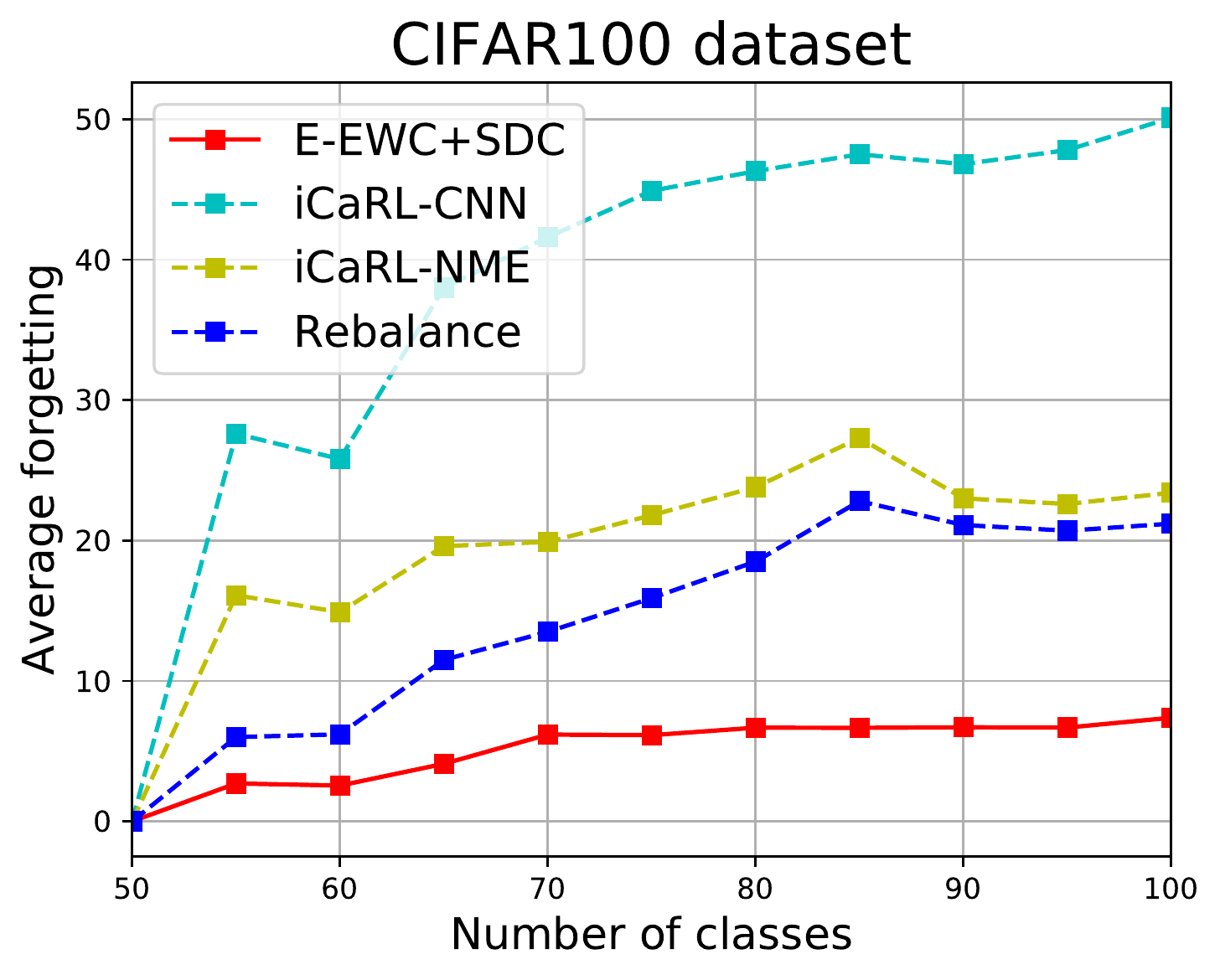}
\includegraphics[width=0.246\textwidth]{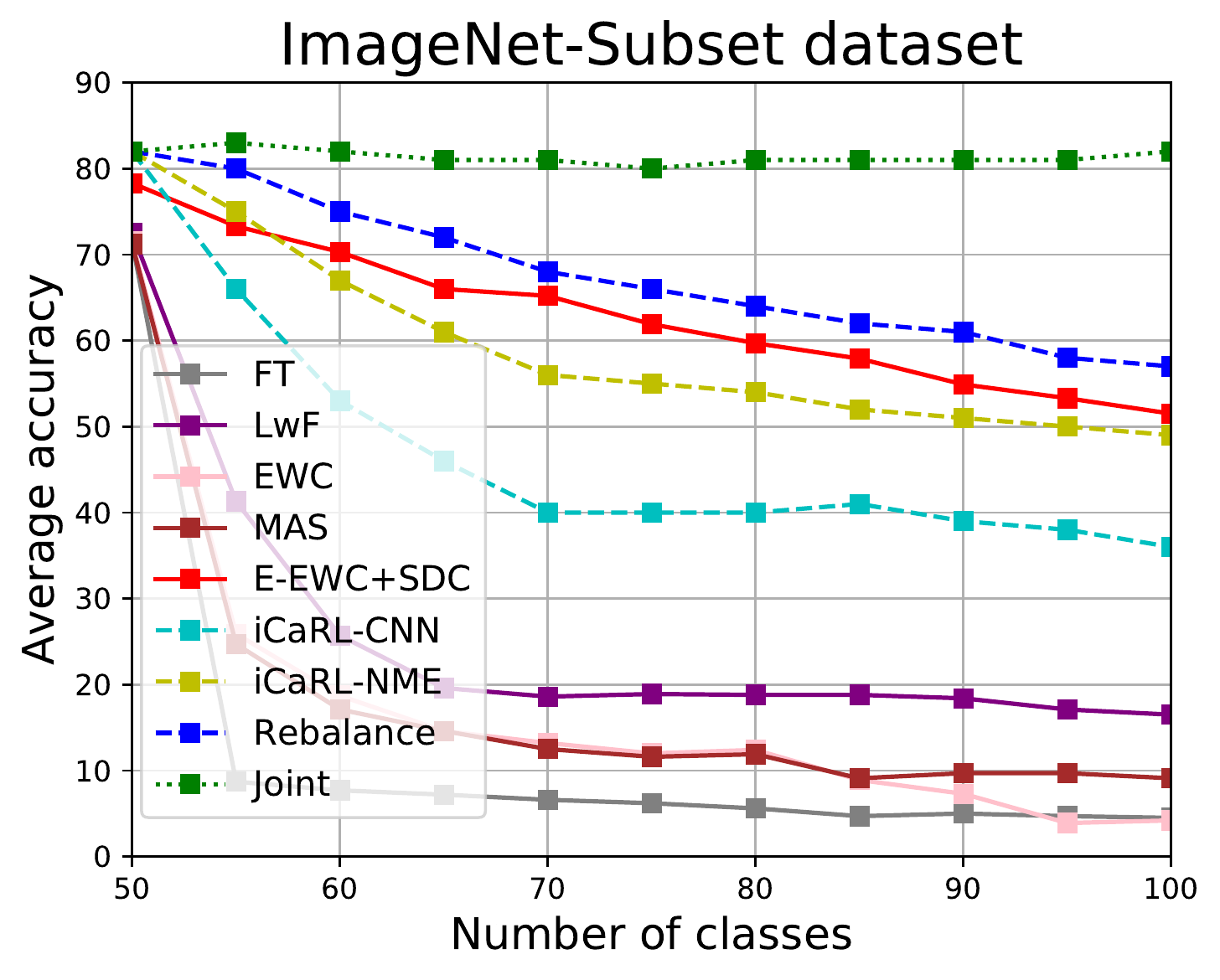}
\includegraphics[width=0.246\textwidth]{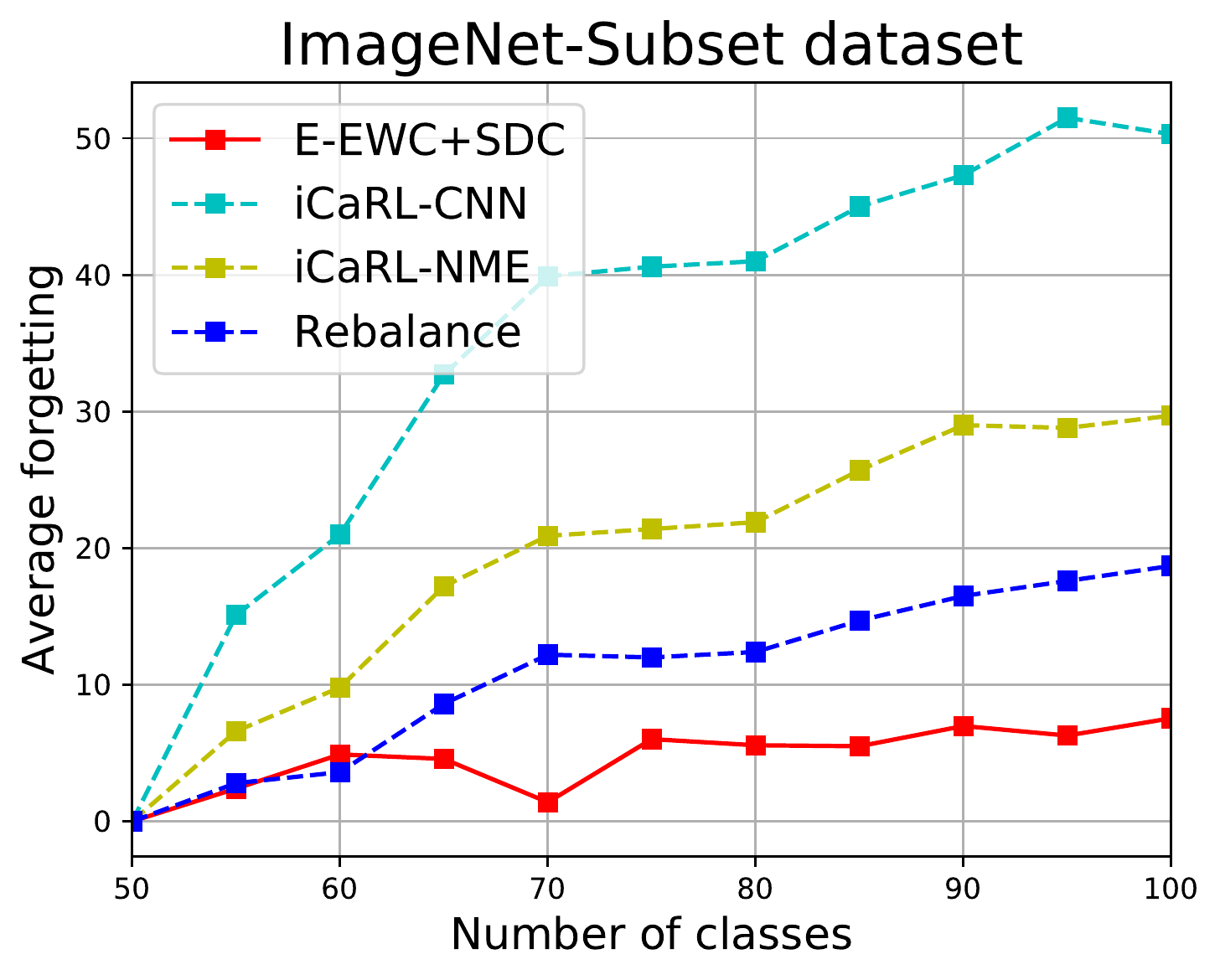}
  \caption{Comparison of average incremental accuracy and average forgetting with eleven-task setting on CIFAR100 and ImageNet-Subset dataset. Solid lines present \textbf{non-exemplar} based methods, dash lines present \textbf{exemplar} based methods.}
  \label{fig:cifar-11}
  \vspace{-1.5\baselineskip}
\end{center}
\end{figure*}


To evaluate the effectiveness of our method, we conduct experiments on two fine-grained datasets: CUB-200-2011 and Flowers-102\footnote{Results on Cars-196 are shown in the supplemental material.} on the six-task scenario. Results are shown in Table~\ref{tab:main}. Here we analyze the average results after training the last task (T6). 

When comparing the various methods to prevent forgetting trained with softmax (LwF/EWC/MAS)  to those applied on embedding network (E-LwF/E-EWC/E-MAS), we observe an enormous gain in performance, showing that embedding networks are less prone to catastrophic forgetting. We also add results for NME on the pre-trained ImageNet model (E-Pre) and the model fixed after training the first task (E-Fix). We can see the best overall accuracy with SDC on two datasets all outperform these two baselines. Furthermore, it can be seen that E-LwF, E-EWC and E-MAS outperform E-FT on both datasets. For example E-EWC obtains a gain of $16.8\%$ on birds and of $50.5\%$ on flowers. The performance of all three methods to prevent forgetting is comparable. Next, we can observe that SDC improves the results of all methods even further, especially for E-FT with $11.9\%$ on birds and $6.2\%$ on flowers. Finally, it is interesting to observe that simple finetuning on embedding networks (E-FT) obtains superior results than LwF, EWC and MAS on birds. When further combined with semantic drift compensation, it further improves over these methods. 

To further analyze if SDC prevents the drift of prototypes, we measure the average distance between the real class-mean and the prototypes (before and after application of SDC). The results are provided in Fig.~\ref{fig:cub-dist-plot}. We observe that SDC reduces the drift of the prototypes. 

\subsection{Comparison to State-of-the-Art Methods}\label{sec:sota} 

\minisection{Ten-task IL on CUB-200 and Caltech-101} To evaluate SDC for longer sequences and compare to Learning without Memorization (LwM), we follow the setting from~\cite{dhar2019learning} and conducted experiments on CUB-200 (100 classes) and Caltech-101, where classes are divided randomly into ten equal tasks. Fig.~\ref{fig:caltech} shows the comparison with FT(softmax), LwM~\cite{dhar2019learning}, Expert Gate~\cite{aljundi2017expert}, the upper bound of joint training, and our best overall methods E-MAS and E-MAS+SDC. We obtain a clear superiority on both datasets. Interestingly, E-MAS already obtains $21.2\%$ and $29.0\%$ higher than the recent LwM method respectively on these two datasets after training 10 tasks. Applying our SDC method further improves the gain further with $6.4\%$ on CUB-200-2011 and $1.4\%$ on Caltech-101. 

\minisection{Experiments on CIFAR100 and ImageNet-Subset} In~\cite{hou2019learning}, the eleven-task evaluation protocol for class-incremental learning was used, where the first task consists of half of the available classes and the rest is split in 10 tasks equally. \emph{Average forgetting} is defined in~\cite{chaudhry2018riemannian} to estimate the forgetting of previous tasks. They quantify forgetting for the $j\mathrm{-th}$ task $f_{j}^{k}=\underset{l\in {1,...,k-1}}{max}(a_{l,j}-a_{k,j}),\forall j<k$, where $a_nm$ is the accuracy of task $n$ after training task $m$. The average forgetting at $k\mathrm{-th}$ task is written as $F_{k} = \frac{1}{k-1}\sum_{j=1}^{k-1}f_{j}^{k}$. 

For the CIFAR100 results of average incremental accuracy and average forgetting are presented in Fig.\ref{fig:cifar-11}. Three groups of methods are shown: non-exemplar based (FT, LwF, EWC, MAS, E-MAS+SDC), exemplar based (iCaRL-CNN~\cite{rebuffi2017icarl}, iCaRL-NME~\cite{rebuffi2017icarl}, Rebalance~\cite{hou2019learning}), and joint training. From the average incremental accuracy, we can see that our overall best method E-EWC+SDC beats all the other non-exemplar based methods by a large margin, with a minimal gap of 27.6\% compared to EWC. It also surpasses two exemplar based methods, namely iCaRL-CNN and iCaRL-NME~\cite{rebuffi2017icarl}, by 7.1\% and 1.1\%. To compare the preventing forgetting capability, we show the performance of our method and exemplar based methods in terms of \emph{average forgetting} metric in Fig.\ref{fig:cifar-11}. Our method (in red) suffers from less forgetting than all the exemplar based methods, obtaining a 13.9\% gain over the best exemplar method (Rebalance~\cite{hou2019learning}). Experiments on ImageNet-Subset outperform all the non-exemplar based methods and two exemplar based methods as well in Fig.\ref{fig:cifar-11}. The conclusion is consistent with CIFAR100 on average incremental accuracy, 35.0\% higher than LwF and 15.5\% higher than iCaRL-CNN and 2.5\% higher than iCaRL-NME. For average forgetting, our method has 3.5\% less forgetting than Rebalance method.

Finally, we also ran fixing the network after finetuning on task one for both datasets. The results are 46.3\% for $CIFAR100$ and 50.5\% on $ImageNet-Sub$. This shows that currently for these difficult multitask settings, methods without exemplars do not significantly outperform this baseline; even some methods with exemplars such as iCaRL-CNN and iCaRL-NME fail on that. This is partially due to the large number of classes in task 1, if we would focus on the performance on the continually learned tasks (2-end) these methods would still report a clear advantage (also see supplementary materials). 

\section{Conclusions}

The dramatic effect of forgetting when applying finetuning, as observed on classification networks, is much less pronounced for embedding networks. This suggest that the current dominance of softmax based methods in continual learning needs to be revisited and our results advocate the usage of embedding networks instead. Furthermore, we proposed a method to approximate the semantic drift of prototypes during training of new tasks. The method is complementary to several existing methods for incremental learning originally designed for classification networks. Experiments show that our method consistently improves results when combined with existing approaches. 

\minisection{Acknowledgement}
We acknowledge the support from Huawei Kirin Solution, the Industrial Doctorate Grant 2016 DI 039 of the Generalitat de Catalunya, the EU Project CybSpeed MSCA-RISE-2017-777720, EU's Horizon 2020 programme under the Marie Sklodowska-Curie grant agreement No.6655919 and the Spanish project RTI2018-102285-A-I00, National Key Laboratory on Blind Signal Processing under grant No.61424131903. 
{\small
\bibliographystyle{ieee}
\bibliography{shortstrings,cvpr}

\begin{thebibliography}{10}\itemsep=-1pt

\bibitem{aljundi2018memory}
R.~Aljundi, F.~Babiloni, M.~Elhoseiny, M.~Rohrbach, and T.~Tuytelaars.
\newblock Memory aware synapses: Learning what (not) to forget.
\newblock In {\em ECCV}, pages 139--154, 2018.

\bibitem{aljundi2017expert}
R.~Aljundi, P.~Chakravarty, and T.~Tuytelaars.
\newblock Expert gate: Lifelong learning with a network of experts.
\newblock In {\em CVPR}, pages 3366--3375, 2017.

\bibitem{belouadah2019il2m}
E.~Belouadah and A.~Popescu.
\newblock Il2m: Class incremental learning with dual memory.
\newblock In {\em Proceedings of the IEEE International Conference on Computer
  Vision}, pages 583--592, 2019.

\bibitem{bromley1994signature}
J.~Bromley, I.~Guyon, Y.~LeCun, E.~S{\"a}ckinger, and R.~Shah.
\newblock Signature verification using a "siamese" time delay neural network.
\newblock In {\em NIPS}, pages 737--744, 1994.

\bibitem{Castro2018}
F.~M. Castro, M.~J. Mar{\'{i}}n-Jim{\'{e}}nez, N.~Guil, C.~Schmid, and
  K.~Alahari.
\newblock {End-to-end incremental learning}.
\newblock {\em Lecture Notes in Computer Science (including subseries Lecture
  Notes in Artificial Intelligence and Lecture Notes in Bioinformatics)}, 11216
  LNCS:241--257, 2018.

\bibitem{chaudhry2018riemannian}
A.~Chaudhry, P.~K. Dokania, T.~Ajanthan, and P.~H. Torr.
\newblock Riemannian walk for incremental learning: Understanding forgetting
  and intransigence.
\newblock In {\em ECCV}, pages 532--547, 2018.

\bibitem{chen2017beyond}
W.~Chen, X.~Chen, J.~Zhang, and K.~Huang.
\newblock Beyond triplet loss: a deep quadruplet network for person
  re-identification.
\newblock In {\em CVPR}, pages 403--412, 2017.

\bibitem{chopra2005learning}
S.~Chopra, R.~Hadsell, and Y.~LeCun.
\newblock Learning a similarity metric discriminatively, with application to
  face verification.
\newblock In {\em CVPR}, volume~1, pages 539--546. IEEE, 2005.

\bibitem{deng2009imagenet}
J.~Deng, W.~Dong, R.~Socher, L.-J. Li, K.~Li, and L.~Fei-Fei.
\newblock Imagenet: A large-scale hierarchical image database.
\newblock In {\em CVPR}, pages 248--255. Ieee, 2009.

\bibitem{dhar2019learning}
P.~Dhar, R.~V. Singh, K.-C. Peng, Z.~Wu, and R.~Chellappa.
\newblock Learning without memorizing.
\newblock In {\em CVPR}, pages 5138--5146, 2019.

\bibitem{fei2004learning}
L.~Fei-Fei, R.~Fergus, and P.~Perona.
\newblock Learning generative visual models from few training examples: An
  incremental bayesian approach tested on 101 object categories.
\newblock In {\em CVPR workshop}, pages 178--178. IEEE, 2004.

\bibitem{he2016deep}
K.~He, X.~Zhang, S.~Ren, and J.~Sun.
\newblock Deep residual learning for image recognition.
\newblock In {\em CVPR}, pages 770--778, 2016.

\bibitem{hoffer2015deep}
E.~Hoffer and N.~Ailon.
\newblock Deep metric learning using triplet network.
\newblock In {\em International Workshop on Similarity-Based Pattern
  Recognition}, pages 84--92. Springer, 2015.

\bibitem{horiguchi2019significance}
S.~Horiguchi, D.~Ikami, and K.~Aizawa.
\newblock Significance of softmax-based features in comparison to distance
  metric learning-based features.
\newblock {\em IEEE Trans. on PAMI}, 2019.

\bibitem{hou2019learning}
S.~Hou, X.~Pan, C.~C. Loy, Z.~Wang, and D.~Lin.
\newblock Learning a unified classifier incrementally via rebalancing.
\newblock In {\em CVPR}, pages 831--839, 2019.

\bibitem{kingma2014adam}
D.~P. Kingma and J.~Ba.
\newblock Adam: A method for stochastic optimization.
\newblock {\em ICLR}, 2014.

\bibitem{kirkpatrick2017overcoming}
J.~Kirkpatrick, R.~Pascanu, N.~Rabinowitz, J.~Veness, G.~Desjardins, A.~A.
  Rusu, K.~Milan, J.~Quan, T.~Ramalho, A.~Grabska-Barwinska, et~al.
\newblock Overcoming catastrophic forgetting in neural networks.
\newblock {\em Proc. Nat. Acad. Sci. USA}, page 201611835, 2017.

\bibitem{krause20133d}
J.~Krause, M.~Stark, J.~Deng, and L.~Fei-Fei.
\newblock 3d object representations for fine-grained categorization.
\newblock In {\em IEEE International Conference on Computer Vision Workshops},
  pages 554--561, 2013.

\bibitem{krizhevsky2009learning}
A.~Krizhevsky, G.~Hinton, et~al.
\newblock Learning multiple layers of features from tiny images.
\newblock Technical report, Citeseer, 2009.

\bibitem{li2018learning}
Z.~Li and D.~Hoiem.
\newblock Learning without forgetting.
\newblock {\em IEEE Trans. on PAMI}, 40(12):2935--2947, 2018.

\bibitem{liu2018rotate}
X.~Liu, M.~Masana, L.~Herranz, J.~Van~de Weijer, A.~M. Lopez, and A.~D.
  Bagdanov.
\newblock Rotate your networks: Better weight consolidation and less
  catastrophic forgetting.
\newblock In {\em Proc. ICPR}, 2018.

\bibitem{lopez2017gradient}
D.~Lopez-Paz and M.~Ranzato.
\newblock Gradient episodic memory for continual learning.
\newblock In {\em NIPS}, pages 6467--6476, 2017.

\bibitem{mallya2018piggyback}
A.~Mallya, D.~Davis, and S.~Lazebnik.
\newblock Piggyback: Adapting a single network to multiple tasks by learning to
  mask weights.
\newblock In {\em ECCV}, pages 67--82, 2018.

\bibitem{mallya2018packnet}
A.~Mallya and S.~Lazebnik.
\newblock Packnet: Adding multiple tasks to a single network by iterative
  pruning.
\newblock In {\em CVPR}, pages 7765--7773, 2018.

\bibitem{masana2018metric}
M.~Masana, I.~Ruiz, J.~Serrat, J.~van~de Weijer, and A.~M. Lopez.
\newblock Metric learning for novelty and anomaly detection.
\newblock In {\em BMVC}, 2018.

\bibitem{masana2020ternary}
M.~Masana, T.~Tuytelaars, and J.~van~de Weijer.
\newblock Ternary feature masks: continual learning without any forgetting.
\newblock {\em arXiv preprint arXiv:2001.08714}, 2020.

\bibitem{mccloskey1989catastrophic}
M.~McCloskey and N.~J. Cohen.
\newblock Catastrophic interference in connectionist networks: The sequential
  learning problem.
\newblock In {\em Psychology of learning and motivation}, volume~24, pages
  109--165. Elsevier, 1989.

\bibitem{mensink2013distance}
T.~Mensink, J.~Verbeek, F.~Perronnin, and G.~Csurka.
\newblock Distance-based image classification: Generalizing to new classes at
  near-zero cost.
\newblock {\em IEEE Trans. on PAMI}, 35(11):2624--2637, 2013.

\bibitem{nilsback2008automated}
M.-E. Nilsback and A.~Zisserman.
\newblock Automated flower classification over a large number of classes.
\newblock In {\em Computer Vision, Graphics and Image Processing}, pages
  722--729. IEEE, 2008.

\bibitem{rajasegaran2019random}
J.~Rajasegaran, M.~Hayat, S.~H. Khan, F.~S. Khan, and L.~Shao.
\newblock Random path selection for continual learning.
\newblock In {\em Advances in Neural Information Processing Systems}, pages
  12648--12658, 2019.

\bibitem{rebuffi2017icarl}
S.-A. Rebuffi, A.~Kolesnikov, G.~Sperl, and C.~H. Lampert.
\newblock icarl: Incremental classifier and representation learning.
\newblock In {\em CVPR}, pages 5533--5542. IEEE, 2017.

\bibitem{rippel2016metric}
O.~Rippel, M.~Paluri, P.~Dollar, and L.~Bourdev.
\newblock Metric learning with adaptive density discrimination.
\newblock In {\em ICLR}, 2016.

\bibitem{rusu2016progressive}
A.~A. Rusu, N.~C. Rabinowitz, G.~Desjardins, H.~Soyer, J.~Kirkpatrick,
  K.~Kavukcuoglu, R.~Pascanu, and R.~Hadsell.
\newblock Progressive neural networks.
\newblock {\em arXiv preprint arXiv:1606.04671}, 2016.

\bibitem{schroff2015facenet}
F.~Schroff, D.~Kalenichenko, and J.~Philbin.
\newblock Facenet: A unified embedding for face recognition and clustering.
\newblock In {\em CVPR}, pages 815--823, 2015.

\bibitem{scott2018adapted}
T.~Scott, K.~Ridgeway, and M.~C. Mozer.
\newblock Adapted deep embeddings: A synthesis of methods for k-shot inductive
  transfer learning.
\newblock In {\em NIPS}, pages 76--85, 2018.

\bibitem{serra2018overcoming}
J.~Serra, D.~Suris, M.~Miron, and A.~Karatzoglou.
\newblock Overcoming catastrophic forgetting with hard attention to the task.
\newblock In {\em ICML}, pages 4555--4564, 2018.

\bibitem{shin2017continual}
H.~Shin, J.~K. Lee, J.~Kim, and J.~Kim.
\newblock Continual learning with deep generative replay.
\newblock In {\em NIPS}, pages 2990--2999, 2017.

\bibitem{simonyan2014very}
K.~Simonyan and A.~Zisserman.
\newblock Very deep convolutional networks for large-scale image recognition.
\newblock {\em arXiv preprint arXiv:1409.1556}, 2014.

\bibitem{snell2017prototypical}
J.~Snell, K.~Swersky, and R.~Zemel.
\newblock Prototypical networks for few-shot learning.
\newblock In {\em NIPS}, pages 4077--4087, 2017.

\bibitem{van2019three}
G.~M. van~de Ven and A.~S. Tolias.
\newblock Three scenarios for continual learning.
\newblock {\em arXiv preprint arXiv:1904.07734}, 2019.

\bibitem{wang2014learning}
J.~Wang, Y.~Song, T.~Leung, C.~Rosenberg, J.~Wang, J.~Philbin, B.~Chen, and
  Y.~Wu.
\newblock Learning fine-grained image similarity with deep ranking.
\newblock In {\em CVPR}, pages 1386--1393, 2014.

\bibitem{wang2017deep}
J.~Wang, F.~Zhou, S.~Wen, X.~Liu, and Y.~Lin.
\newblock Deep metric learning with angular loss.
\newblock In {\em ICCV}, pages 2612--2620. IEEE, 2017.

\bibitem{Wang2019}
X.~Wang, X.~Han, W.~Huang, D.~Dong, and M.~R. Scott.
\newblock Multi-similarity loss with general pair weighting for deep metric
  learning.
\newblock In {\em CVPR}, pages 5022--5030, 2019.

\bibitem{WelinderEtal2010}
P.~Welinder, S.~Branson, T.~Mita, C.~Wah, F.~Schroff, S.~Belongie, and
  P.~Perona.
\newblock {Caltech-UCSD Birds 200}.
\newblock Technical Report CNS-TR-2010-001, California Institute of Technology,
  2010.

\bibitem{wu2018memory}
C.~Wu, L.~Herranz, X.~Liu, Y.~Wang, J.~van~de Weijer, and B.~Raducanu.
\newblock Memory replay gans: learning to generate images from new categories
  without forgetting.
\newblock In {\em NIPS}, 2018.

\bibitem{wu2019large}
Y.~Wu, Y.~Chen, L.~Wang, Y.~Ye, Z.~Liu, Y.~Guo, and Y.~Fu.
\newblock Large scale incremental learning.
\newblock In {\em CVPR}, 2019.

\bibitem{yang2018robust}
H.-M. Yang, X.-Y. Zhang, F.~Yin, and C.-L. Liu.
\newblock Robust classification with convolutional prototype learning.
\newblock In {\em CVPR}, pages 3474--3482, 2018.

\bibitem{yu2019learning}
L.~Yu, V.~O. Yazici, X.~Liu, J.~Van~de Weijer, Y.~Cheng, and A.~Ramisa.
\newblock Learning metrics from teachers: Compact networks for image embedding.
\newblock In {\em Proceedings of the IEEE Conference on Computer Vision and
  Pattern Recognition}, pages 2907--2916, 2019.

\bibitem{zenke2017continual}
F.~Zenke, B.~Poole, and S.~Ganguli.
\newblock Continual learning through synaptic intelligence.
\newblock In {\em ICML}, pages 3987--3995. JMLR. org, 2017.

\end{thebibliography}
}

\clearpage

\begin{appendices}

\section{Visualization of E-LwF and E-MAS}
\begin{figure*}[tb]
\centering
\subfigure[]{\includegraphics[width=0.3\textwidth]{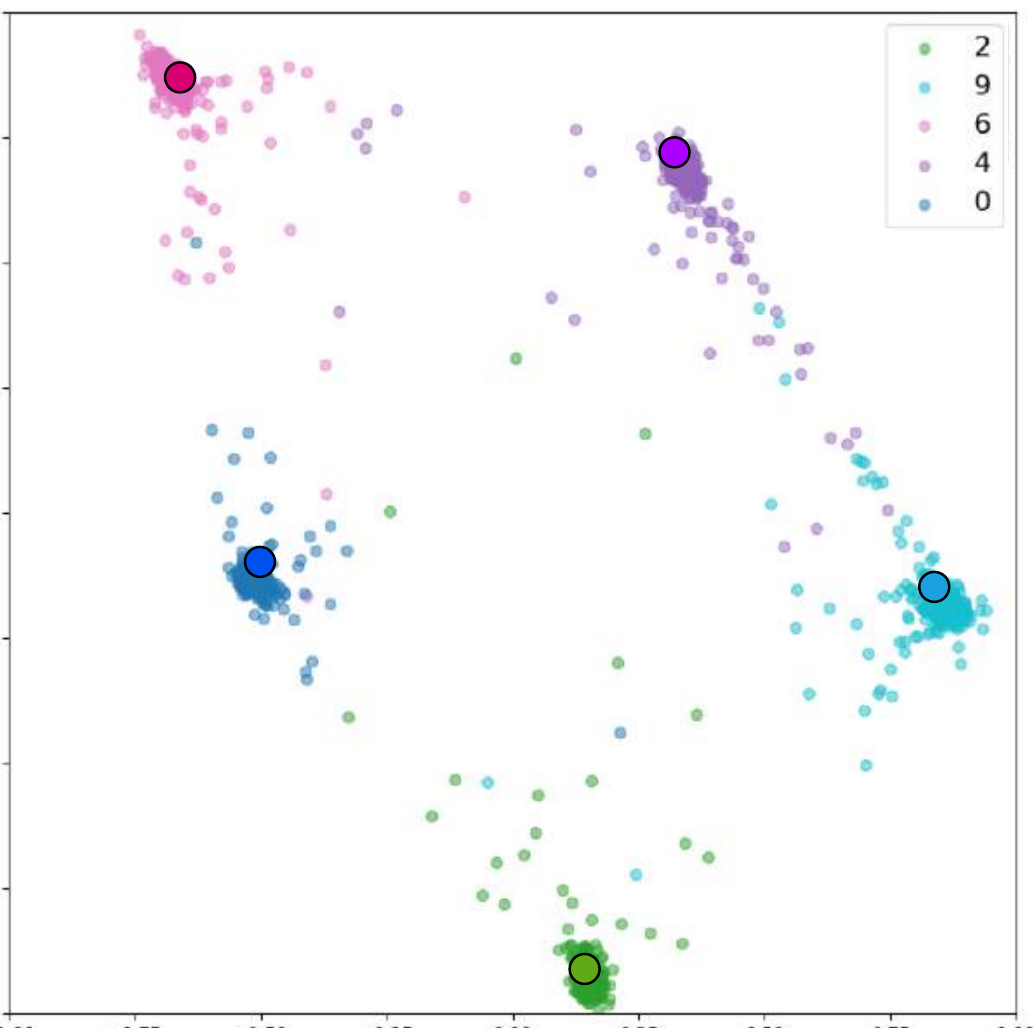}}
\subfigure[]{\includegraphics[width=0.3\textwidth]{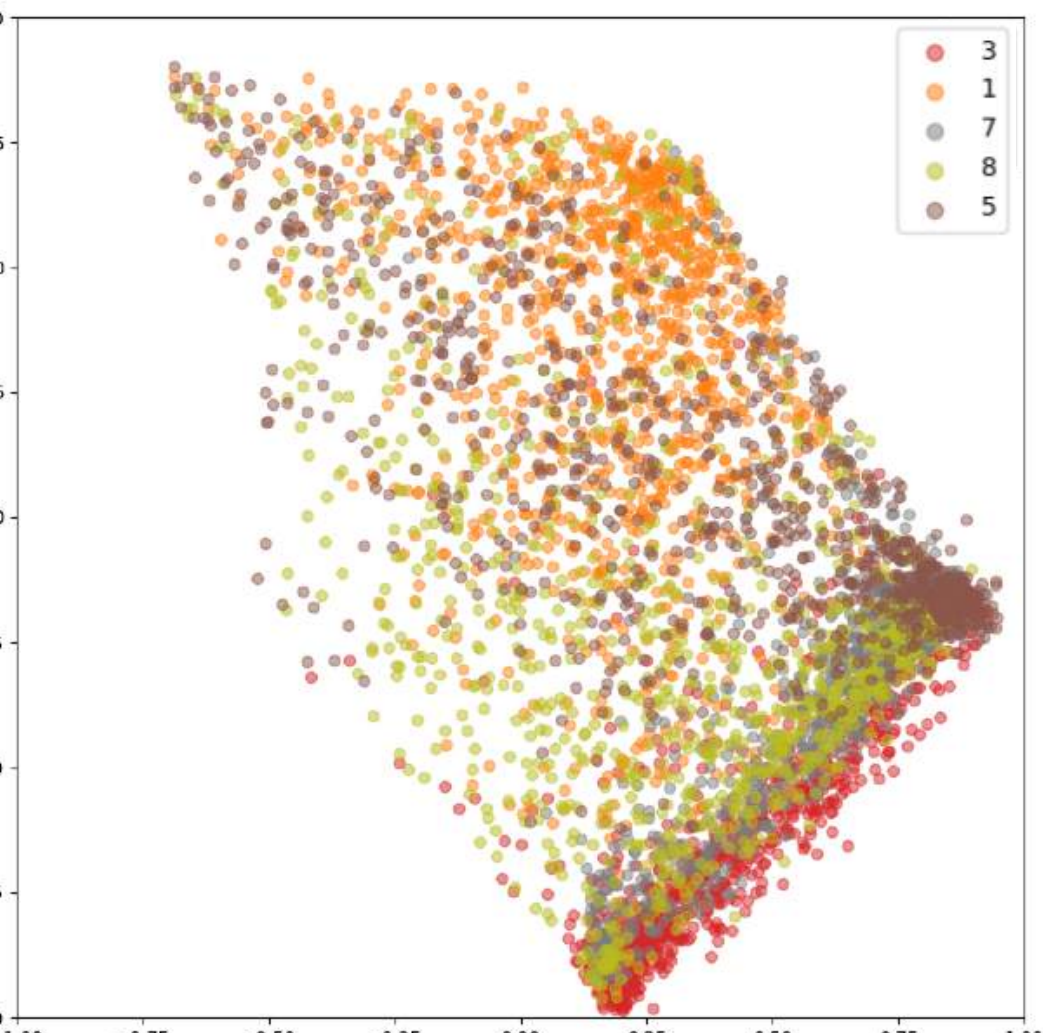}}
\subfigure[]{\includegraphics[width=0.3\textwidth]{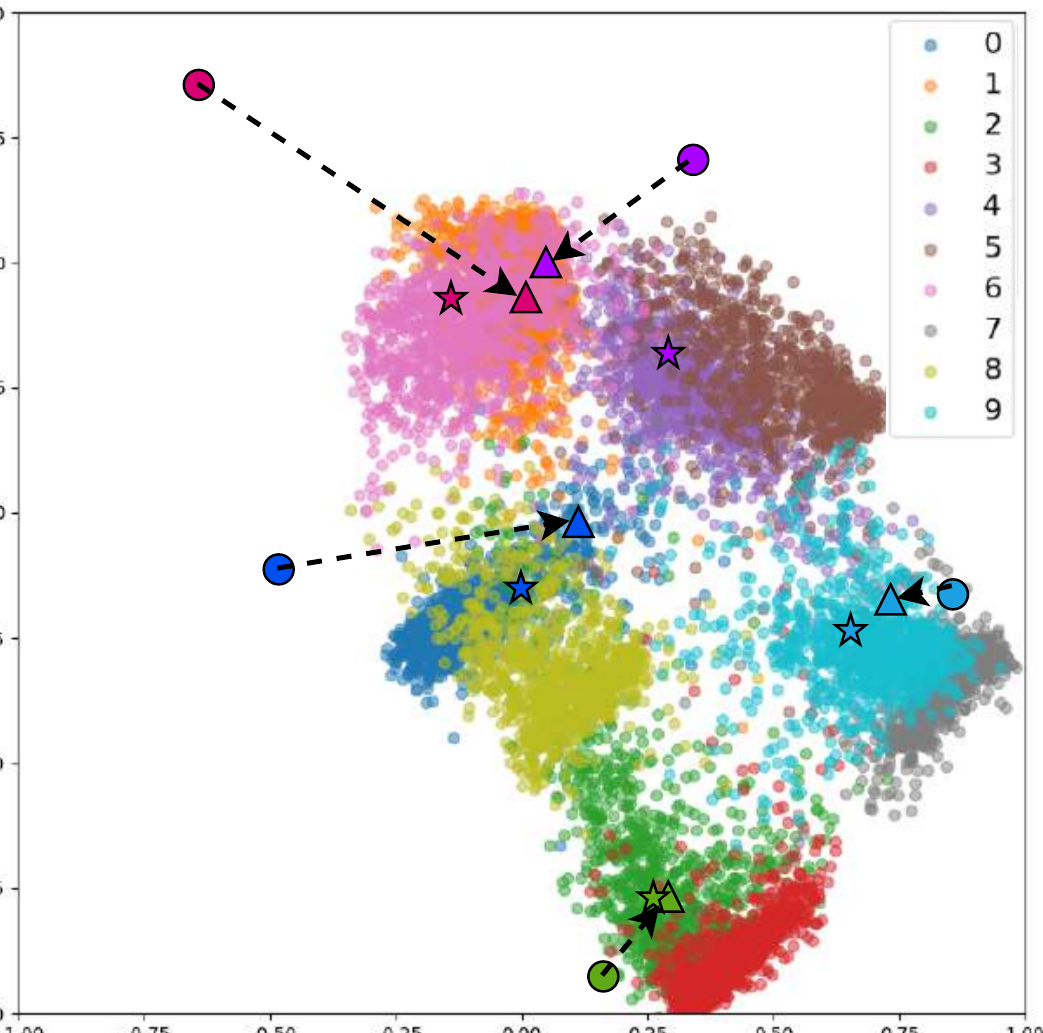}} \\
\subfigure[]{\includegraphics[width=0.3\textwidth]{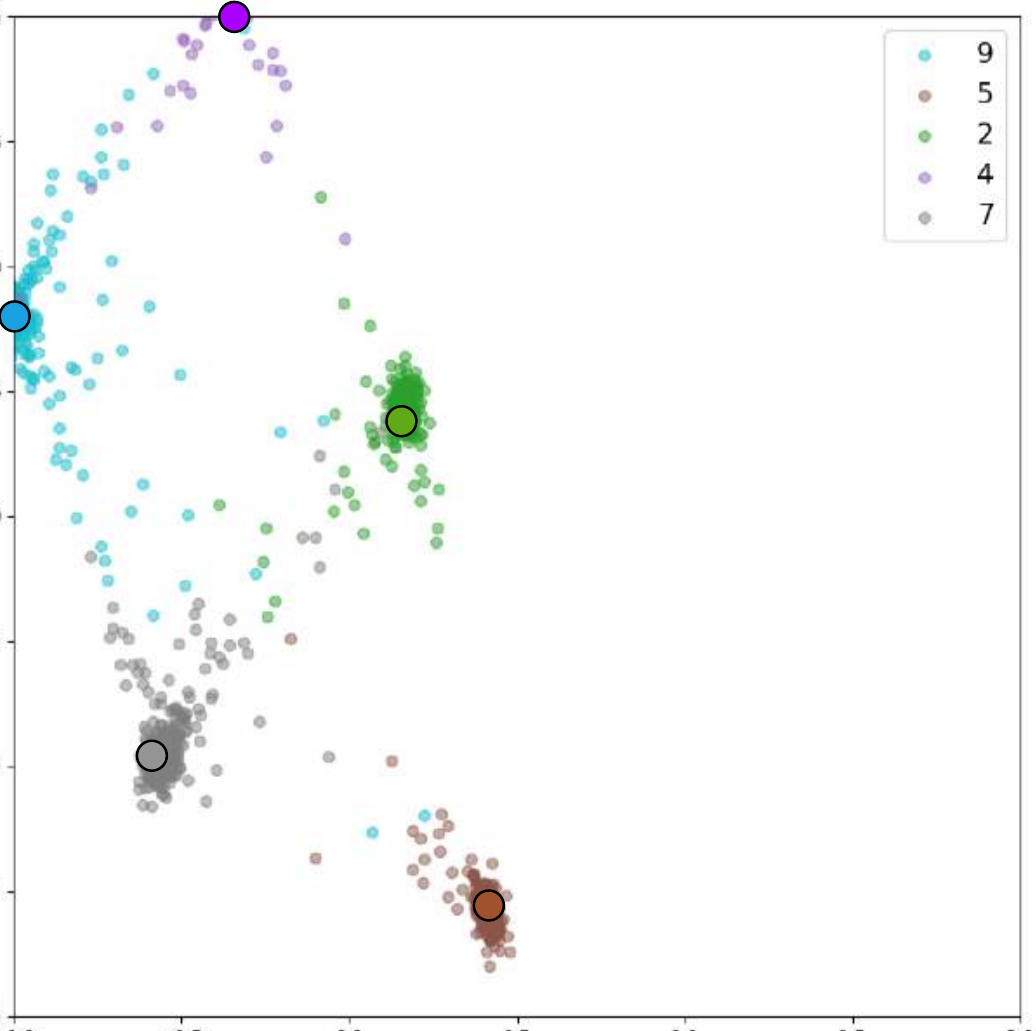}}
\subfigure[]{\includegraphics[width=0.3\textwidth]{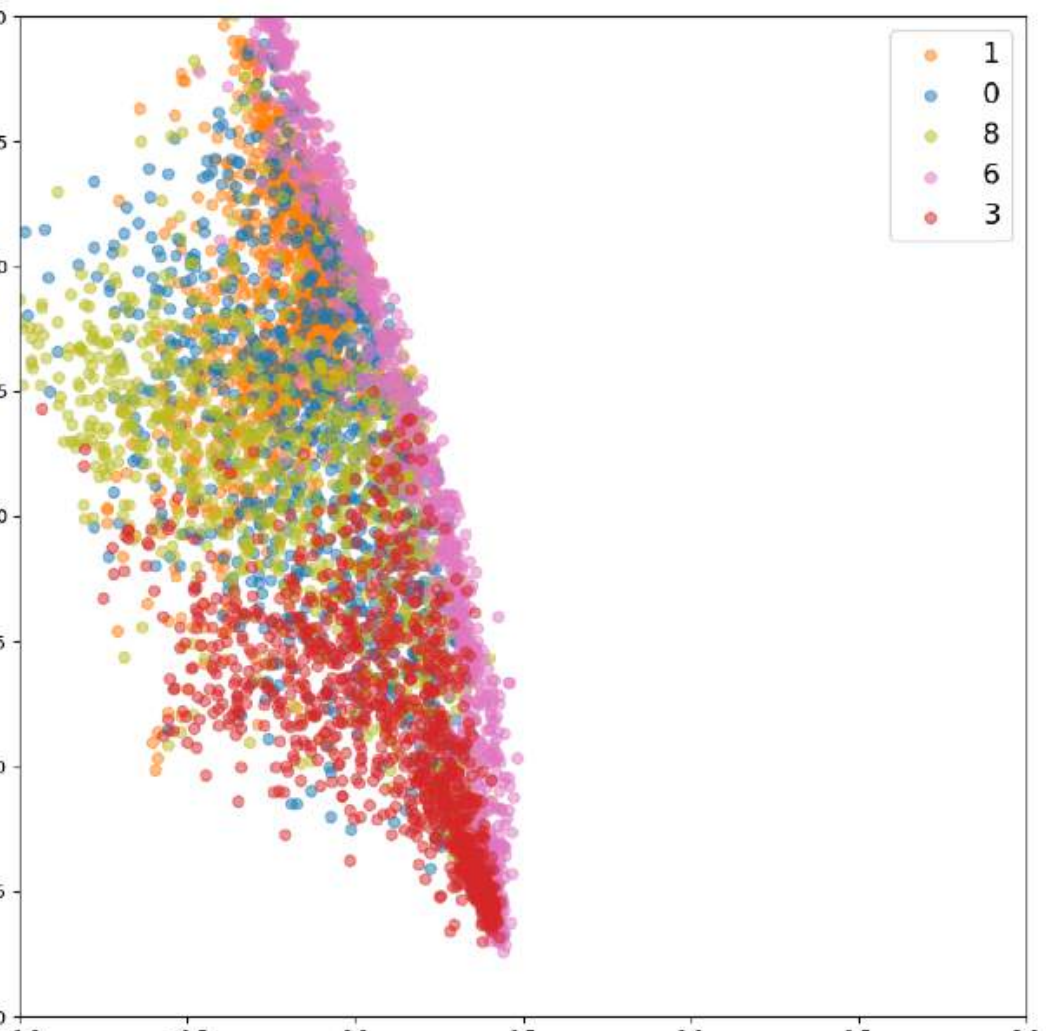}}
\subfigure[]{\includegraphics[width=0.3\textwidth]{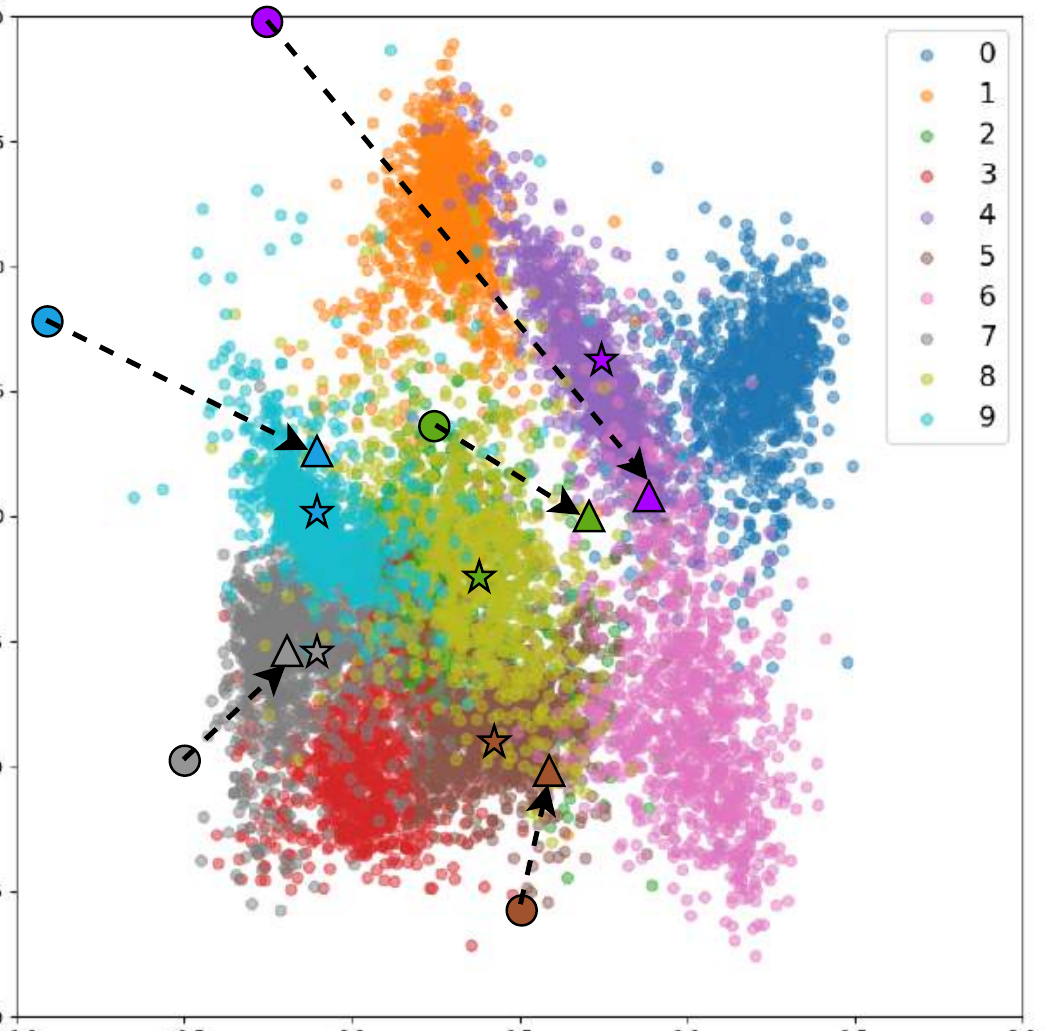}} \\
\caption{Examples of the drift vectors in the cases of E-LWF (top) and E-MAS (bottom). (a) and (d) represent the embedding of 5 classes of task 1 after training task 1; (b) and (e) represent the embedding of another 5 classes of task 2 after training task 1; (c) and (f) show the embeddings of 10 classes of two tasks together. The saved prototypes of the previous task(indicated by round) are estimated to new positions (indicated by triangle) by our proposed SDC in the new model which is observed to be closer to the real mean (indicated by star). The dotted arrows are the SDC vectors.} 
\label{fig:real DC2}
\end{figure*}

In  Fig.~\ref{fig:real DC2} we show examples of the drift vectors which are estimated by SDC in the case of E-LwF and E-MAS to supplement Fig.~4 in the main paper.

\section{Pre-trained Model without Birds}

The results presented in Table.~1 in the main paper are based on ResNet18 pre-trained from the ImageNet dataset. As some of the categories of birds are present in the ImageNet, we conducted additional experiments. We removed 59 classes in total form the original dataset, including birds (e.g.~macaw, flamingo, black swan) and similar species (e.g.~cock, hen, king penguin), and then trained ResNet18 network on the constrained dataset. Table.~\ref{table:no_birds} shows the average incremental accuracy results pre-trained from ImageNet without birds. It can be seen that there is no significant difference with fine-tuning, while slightly worse with the other three methods compared to Table.~1, where bird categories were used in the pre-trained network.

\section{Results with Multi-similarity Loss and Angular Loss }

A triplet loss is used in the main paper as a default metric loss function. Additionally, we investigated two newer versions of metric losses: Multi-similarity~\cite{Wang2019} and Angular loss~\cite{wang2017deep} on CUB-200-2011 with a class-IL setting. The results are shown in Table.~\ref{table:loss}. We can see that the accuracy for the first task with the angular loss is 5.0\% lower than for the triplet loss, while the multi-similarity starts with 4.0\% higher accuracy. For E-FT method, a multi-similarity loss can achieve much better average incremental accuracy after training six tasks with a 13.3\% improvement compared to the triplet loss. It is interesting to note that after adding our SDC, it achieves 56.1\% after the final task, which is even better than other methods with regularization and SDC except for E-EWC+SDC. For the angular loss E-FT and E-FT+SDC present slightly lower results in comparison to the others regularized and regularized with SDC methods. Despite addressing some of the triplet loss function shortcomings, both of new losses obtain similar results for class-IL to the triplet loss used for all experiments in the main paper.

\setcounter{table}{1}
\begin{table}[tb]
\centering
\caption{Average incremental accuracy for CUB-200-2011 datasets with constrained pre-trained ImageNet.}
\resizebox{\linewidth}{!}{%
\begin{tabular}{c|cccccc}
\hline
\begin{tabular}[c]{@{}c@{}}Pre-trained ImageNet\\ (w/o birds)\end{tabular} & T1 & T2 & T3 & T4 & T5 & T6\\ \hline
FT & 79.1 & 33.5 & 23.2 & 17.3 & 14.3 & 10.0 \\ \hline
E-FT & 86.3 & 74.6 & 63.2 & 54.8 & 43.8 & 37.5 \\ \hline
LwF & 79.1 & 51.7 & 37.0 & 28.7 & 24.8 & 19.5 \\ \hline
E-LwF &  86.3 & 76.4 & 67.7 & 60.1 & 55.7 & 50.8 \\ \hline
EWC & 79.1 & 37.8 & 27.3 & 18.0 & 14.6 & 10.2 \\ \hline
E-EWC & 86.3 & 73.9 & 63.2 & 59.0 & 53.4 & 50.7\\ \hline
MAS & 79.1 & 44.5 & 32.1 & 27.2 & 23.2 & 19.4 \\ \hline
E-MAS & 86.3 & 73.2 & 61.1 & 55.9 & 51.1 & 48.6 \\ \hline
\end{tabular}}
\label{table:no_birds}
\end{table}

\begin{table*}[tb]
\centering
\caption{Average incremental accuracy for CUB-200-2011 datasets with Multi-similarity and Angular loss loss.}\label{table:loss}
\begin{tabular}{c|cccccc|cccccc}
\hline
 & \multicolumn{6}{c|}{Multi-similarity} & \multicolumn{6}{c}{Angular} \\
 & T1 & T2 & T3 & T4 & T5 & T6 & T1 & T2 & T3 & T4 & T5 & T6\\ \hline
E-FT &88.1&74.4&65.5&59.8&52.2&50.7&79.1&61.7&50.9&48.1&40.9&40.5 \\ \hline
E-FT+SDC &88.1&76.4&69.9&63.0&59.5&56.1&79.1&65.8&57.6&53.4& 49.6&45.5\\ \hline
E-LwF &88.1&74.3&66.5&61.6&56.6&50.9&79.1&70.8&61.9&56.0&50.9&45.4 \\ \hline
E-LwF+SDC &88.1&74.7&66.9&61.3&57.4&51.5&79.1&69.6&60.6&55.5&51.2&46.6 \\ \hline
E-EWC &88.1&75.2&66.3&62.0&55.2&52.9&79.1&66.3&57.5&53.4&48.3&44.6\\ \hline
E-EWC+SDC &88.1&76.5&67.9&64.0&60.4&57.7&79.1&67.7&59.9&55.5& 51.2&48.6\\ \hline
E-MAS &88.1&74.9&64.9&59.9&54.1&51.2&79.1&68.0&59.1&54.1&46.4&46.4 \\ \hline
E-MAS+SDC &88.1&76.1&66.8&63.0&58.6&55.7&79.1&67.9&60.8&56.5&52.0&49.0 \\ \hline
\end{tabular}
\end{table*}

\section{Confusion Matrix}

\begin{figure*}[tb]
\centering
\subfigure[]{\includegraphics[width=0.4\textwidth]{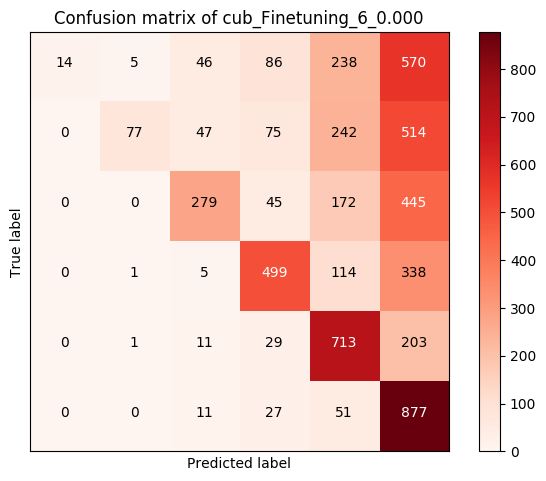}}
\subfigure[]{\includegraphics[width=0.4\textwidth]{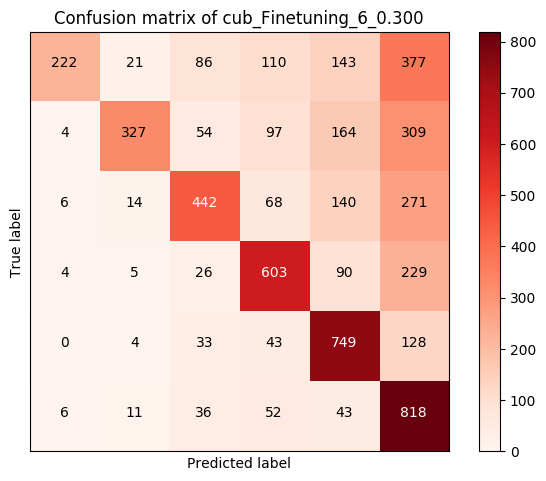}} \\
\subfigure[]{\includegraphics[width=0.4\textwidth]{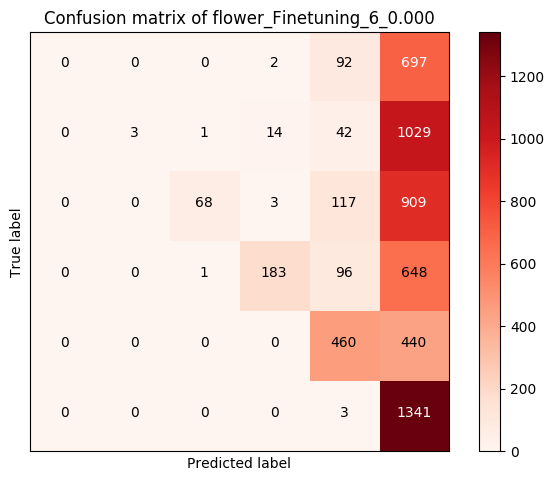}}
\subfigure[]{\includegraphics[width=0.4\textwidth]{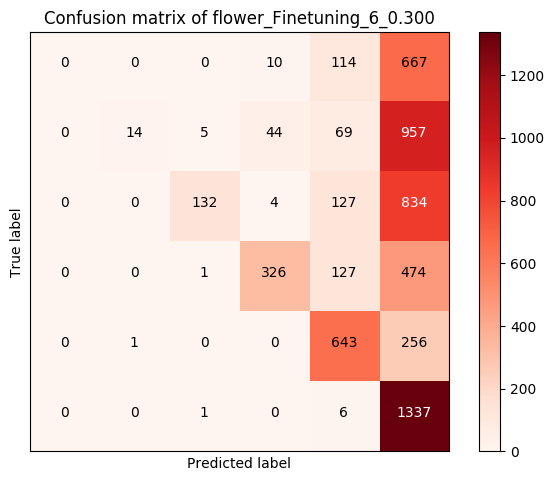}} \\
\caption{Confusion matrix of CUB-200-2011 and Flowers-102 with Fine-tuning method before applying SDC (a, c) and after applying SDC (b, d). } 
\label{fig:cm}
\end{figure*}

We show confusion matrix of CUB-200-2011 and Flowers-102 dataset with Fine-tuning respectively in Fig.~\ref{fig:cm}, for further insight of our SDC method. The left figures are the confusion matrices before applying SDC, the right ones are after applying SDC. We can see that our SDC method is able to compensate the forgetting of the previous tasks to some extent.

\section{Experiments on VGG} 
\begin{figure}[tb]
\begin{center}
  \includegraphics[width=0.45\textwidth]{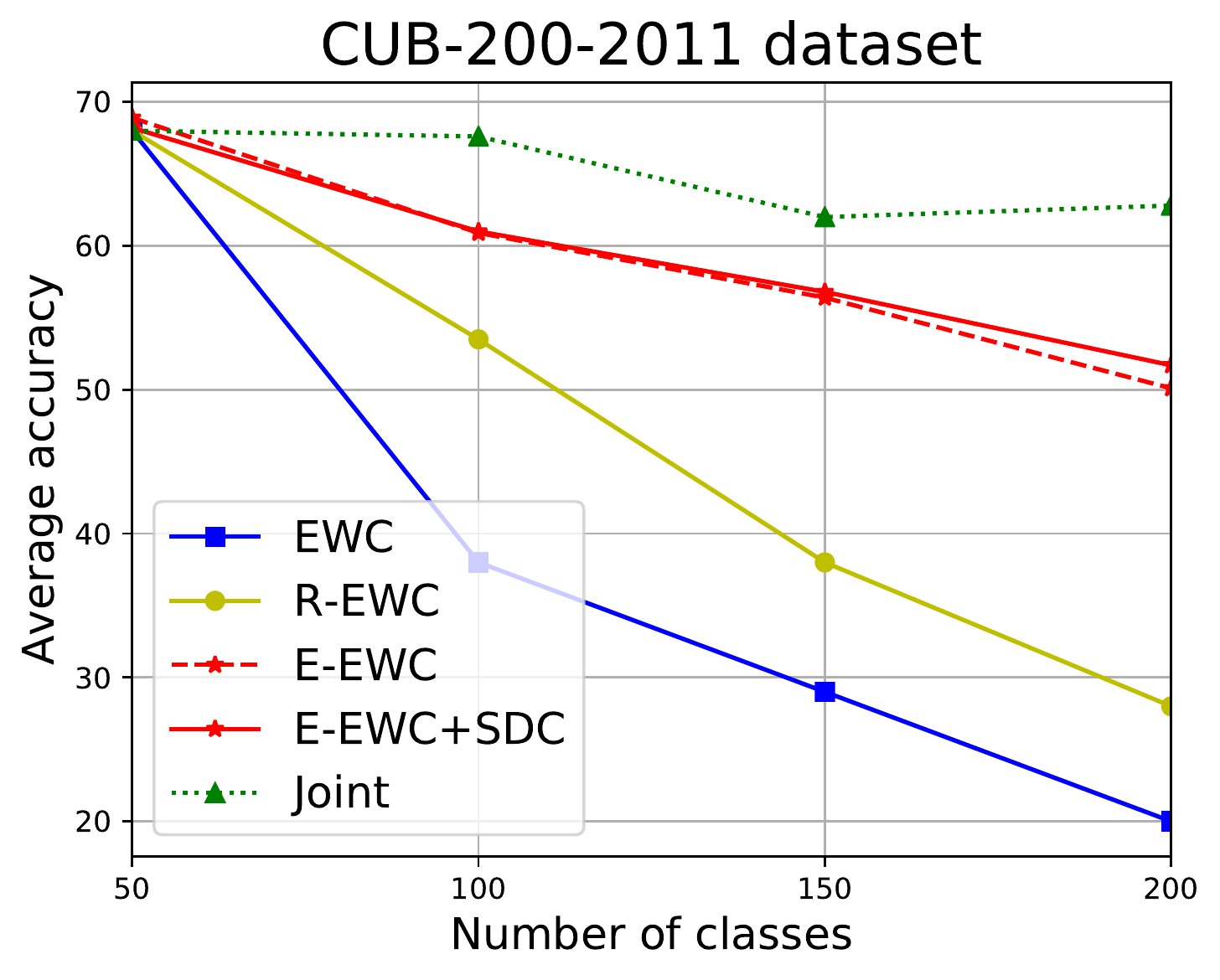}
  \caption{Comparison of four-task with VGG16 network.}
  \label{fig:vgg}
\end{center}
\end{figure}

To be able to compare our method to R-EWC and validate its generalization ability, we follow the protocol of Liu et al.~\cite{liu2018rotate}  and implement our method on a VGG16~\cite{simonyan2014very}.
The CUB-200 dataset is divided into four equal tasks; the same setting as in Table.1. The comparison of different methods is shown in Fig.~\ref{fig:vgg}. We can see that our E-EWC surpasses EWC~\cite{kirkpatrick2017overcoming} and R-EWC~\cite{liu2018rotate} with clear superiority, improving with 30.1\% and 22.1\% respectively. SDC contributes an additional 1.6\% gain.

\section{Classification with Embedding Networks on Cars-196 Dataset}

Cars-196 dataset~\cite{krause20133d} contains $16,185$ images of $196$ cars classes. ResNet-18~\cite{he2016deep} is adopted as the backbone network pretrained from ImageNet for Cars-196 dataset as well. We train our model with learning rate $1\mathrm{e}{-5}$ for $100$ epochs on cars, the other settings are the same as birds and flowers. Results are shown in Table.~\ref{tab:car} after training the last task (T6). The conclusion is consistent with the CUB-200 and Flowers-102 datasets.

\begin{table}[tb]
    \centering
    \caption{Average incremental accuracy for Cars-196 dataset.}\label{tab:car}
    \begin{tabular}{c|cccccc}
    \hline
    \multicolumn{1}{l|}{} &  \multicolumn{6}{c}{Cars-196}\\
     & T1 & T2 & T3 & T4 & T5 & T6 \\
    \hline
    \footnotesize{E-Pre} & 44.0 & 34.5 & 27.4 & 24.8 & 23.5& 22.3\\
    \footnotesize{E-Fix} & 58.2 & 45.9 & 38.6 & 33.8 &32.1 & 30.5\\
    \hline
    \footnotesize{FT} & 67.5 & 33.0 & 24.2 & 19.6 &15.0 &13.6 \\
    \footnotesize{E-FT} & 58.2 & 44.8 &34.7  &30.2 &23.6 & 17.3\\
    \footnotesize{E-FT+SDC} & 58.2 &  \textbf{50.3} &  \textbf{41.8} & \textbf{34.0} &  \textbf{25.4} & \textbf{18.2} \\
    \hline
    \footnotesize{LwF} & 67.5 & 40.3 & 33.3 & 30.1 &26.7 & 21.9\\
    \footnotesize{E-LwF} & 58.2& 48.2&40.9 &36.2  & 34.2 &32.0 \\
    \footnotesize{E-LwF+SDC} & 58.2 & \textbf{47.2} &  \textbf{41.8} & \textbf{36.8} & \textbf{35.4} & \textbf{33.9} \\
    \hline
    \footnotesize{EWC} & 67.5 & 30.8 & 25.8&19.9 &16.5 &15.6 \\
    \footnotesize{E-EWC} &58.2  &  47.0& 39.6 &35.1 & 32.9 & 30.7\\
    \footnotesize{E-EWC+SDC} & 58.2 & \textbf{48.1} &  \textbf{40.9} &\textbf{36.4} & \textbf{34.0} & \textbf{32.2} \\
    \hline
    \footnotesize{MAS} & 67.5 & 37.1 & 27.7& 22.9&20.2 &17.0 \\
    \footnotesize{E-MAS} & 58.2 &  \textbf{46.3} &38.3  & 33.6 & 31.4 & 28.8 \\
    \footnotesize{E-MAS+SDC} & 58.2 & \textbf{46.3} &  \textbf{39.0} & \textbf{34.0} & \textbf{31.8} & \textbf{30.7} \\
    \bottomrule
    \end{tabular}
\end{table}
    
\section{Experiments on CIFAR100 and ImageNet-Subset}

We show the details of the average accuracy of our methods on CIFAR100 and ImageNet-Subset followed by the eleven-task evaluation protocol~\cite{hou2019learning} in Table~\ref{tab:cifar} (E-EWC+SDC is shown in Fig.~7 in the main paper). Batch nomalization is fixed after training the first task. It can be seen that E-LwF, E-EWC and E-MAS outperform E-FT on both datasets. Also we can observe that SDC improves the results of all methods even further except for E-LwF, especially for E-FT with $7.4\%$ on CIFAR100, and $3.5\%$ on ImageNet-Subset. Essentially, E-EWC and E-MAS indirectly limit the drift of the embedding by constraining the important weights, whereas E-LwF is directly constraining the embedding, which in the end results in less drift. 

\begin{figure}[tb]
\begin{center}
    \includegraphics[width=0.4\textwidth]{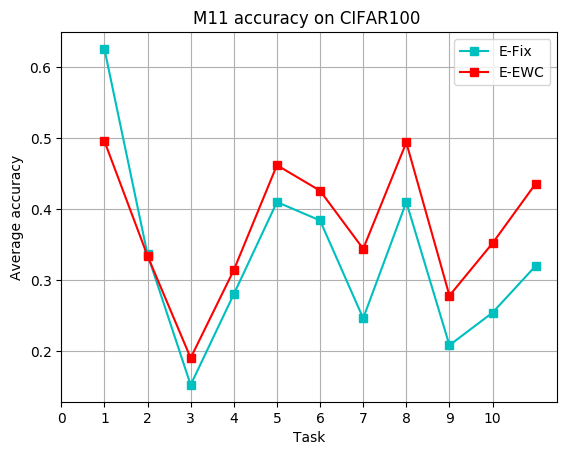}
    \includegraphics[width=0.4\textwidth]{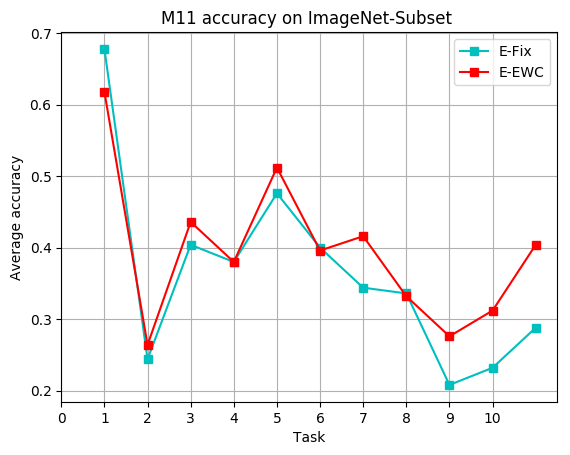}
  \caption{Accuracy of each of the eleven tasks with E-Fix and after training all tasks with E-EWC on CIFAR100 and ImageNet-Subset dataset. }
  \label{fig:cifar2}
\end{center}
\end{figure}

As discussed in the main paper, the good results of E-Fix for these more difficult datasets shows that continual learning methods without exemplars have difficulty outperforming this baseline (and even some methods which use exemplars like iCaRL). In Fig.~\ref{fig:cifar2} we also show the accuracy of each task after training the eleven tasks for E-Fix (in cyan) and E-EWC (in red). We can see that E-EWC always outperforms E-Fix except for the first task. It means even though the average accuracy of the eleven tasks with E-Fix and E-EWC is similar, freezing the first model does not have any positive forward transfer.

\begin{table}[tb]
    \centering
    \caption{Average incremental accuracy for CIFAR100 and ImageNet-Subset.}\label{tab:cifar}
    \begin{tabular}{c|c|c}
    \toprule
    \multicolumn{1}{l|}{} & \multicolumn{1}{c|}{CIFAR100} & 
    \multicolumn{1}{c}{ImageNet-Subset} \\
     & T11 & T11\\
    \hline
    E-Fix & 46.3 & 50.5\\
    \hline
    E-FT & 37.4 & 47.4\\
    E-FT+SDC & \textbf{44.8} & \textbf{50.9}\\
    \hline
    E-LwF & 46.1 & \textbf{51.5}\\
    E-LwF+SDC & 46.1 & 50.5\\
    \hline
    E-EWC & 40.8& 49.5\\
    E-EWC+SDC & \textbf{46.1}& \textbf{51.5}\\
    \hline
    E-MAS & 43.1& 50.8\\
    E-MAS+SDC & \textbf{46.3}&\textbf{51.2}\\
    \bottomrule
    \end{tabular}
\end{table}

\end{appendices}

\end{document}